\documentclass[11pt]{article}

\usepackage[final]{acl}

\usepackage{times}
\usepackage{latexsym}

\usepackage[T1]{fontenc}
\usepackage{amssymb}
\usepackage{amsmath}
\usepackage{multirow}
\usepackage{soul}
\usepackage[table]{xcolor}

\usepackage{booktabs}
\usepackage{longtable}
\usepackage{array}
\usepackage{tcolorbox}
\tcbuselibrary{skins,breakable}
\usepackage{enumitem}
\usepackage{textcomp}
\usepackage{algorithm}
\usepackage{algorithmic}
\usepackage{tabularx}

\hypersetup{colorlinks=true, linkcolor=blue, citecolor=blue, urlcolor=blue}

\newcommand{\red}[1]{\textcolor{red!70!black}{#1}}

\newcommand{\brown}[1]{\textcolor{brown!80!black}{#1}}
\newcommand{\shade}[1]{\colorbox{yellow!25}{#1}}

\usepackage[utf8]{inputenc}

\usepackage{microtype}

\usepackage{inconsolata}

\usepackage{graphicx}
\usepackage{xspace}
\newcommand{\our}{\textsc{MedCoast}\xspace}

\title{\our: Semantic Abstraction of Reasoning Trajectories for Test-Time Biomedical Question Answering}

\author{
  Yishan Zhong$^{\spadesuit}$,
  Jieru Li$^{\spadesuit}$,
  Xuanzhe Ding$^{\spadesuit}$,
  Shaunna Wang$^{\spadesuit}$,
  Yanxin Yang$^{\spadesuit}$,\\
  \bf Micky C. Nnamdi$^{\spadesuit}$,
  J. Ben Tamo$^{\spadesuit}$,
  Yifei Wang$^{\spadesuit}$,
  Yining Yuan$^{\spadesuit}$,
  Benoit L. Marteau$^{\spadesuit}$,\\
  \bf Shaun Q. Y. Tan$^{\spadesuit}$,
  Carrie Moore$^{\heartsuit}$,
  Luyu Xie$^{\heartsuit}$,
  Yang Xie$^{\heartsuit}$,
  Wenqi Shi$^{\heartsuit\clubsuit}$,
  \\
  \textbf{and May D. Wang$^{\spadesuit}$}
  \\
  $^{\spadesuit}$Georgia Institute of Technology \qquad
  $^{\heartsuit}$University of Texas Southwestern Medical Center
  \\
  $^{\clubsuit}$Washington University in St.\ Louis
  \\
  Corresponding authors: \texttt{maywang@gatech.edu}
}

\begin{document}
\maketitle

\begin{abstract}
Biomedical question answering often requires multi-step reasoning under limited inference budgets. While test-time search methods such as Tree-of-Thought can improve reasoning performance, directly searching over free-form language is computationally expensive and highly redundant, as many reasoning trajectories contain semantically equivalent intermediate steps expressed in different surface forms. We introduce \our, an offline-to-online reasoning framework that converts sampled chain-of-thought trajectories into a compact space for structured process supervision. Offline, we recursively cluster semantically similar reasoning steps under shared prefixes to construct an abstract reasoning tree that captures recurring clinical reasoning patterns. We then derive node-level supervision from progress toward correct solutions, and train a lightweight process reward model (PRM) to evaluate partial reasoning states. At inference time, the learned critic guides test-time exploration by prioritizing promising reasoning prefixes. Across seven biomedical QA benchmarks, \our often improves over self-consistency under fixed inference budgets, with gains of up to 15 points. Despite using a reward model more than 50 times smaller, \our achieves performance comparable to state-of-the-art reward models in several settings, highlighting its potential for resource-constrained biomedical reasoning. 
\footnote{Details of our implementation are available at: \href{https://github.com/zhongys97/MedCoast}{https://github.com/zhongys97/MedCoast}. The weights of trained PRMs are downloadable from \href{https://huggingface.co/zhongys97/MedCoast}{https://huggingface.co/zhongys97/MedCoast}.}
\end{abstract}

\section{Introduction}

Biomedical question answering remains a challenging setting for large language models (LLMs), as questions often require multi-step clinical reasoning, specialized terminology, and the correct application of guidelines, mechanisms, or pathophysiology \cite{jin2021disease, xu2025medagentgym}. Chain-of-thought (CoT) prompting improves performance by encouraging intermediate reasoning \cite{wei2022chain}, but medical reasoning remains brittle under distribution shift and the constrained inference budgets typical of practical deployment \cite{shi2024medadapter}. This challenge is inherently sequential: early reasoning decisions shape later conclusions, multiple plausible hypotheses may coexist before evidence resolves ambiguity, and the correct final answer may not be identifiable from the first few steps \cite{yao2023tree, besta2024graph}.

\begin{figure}
    \centering
    \includegraphics[width=\linewidth]{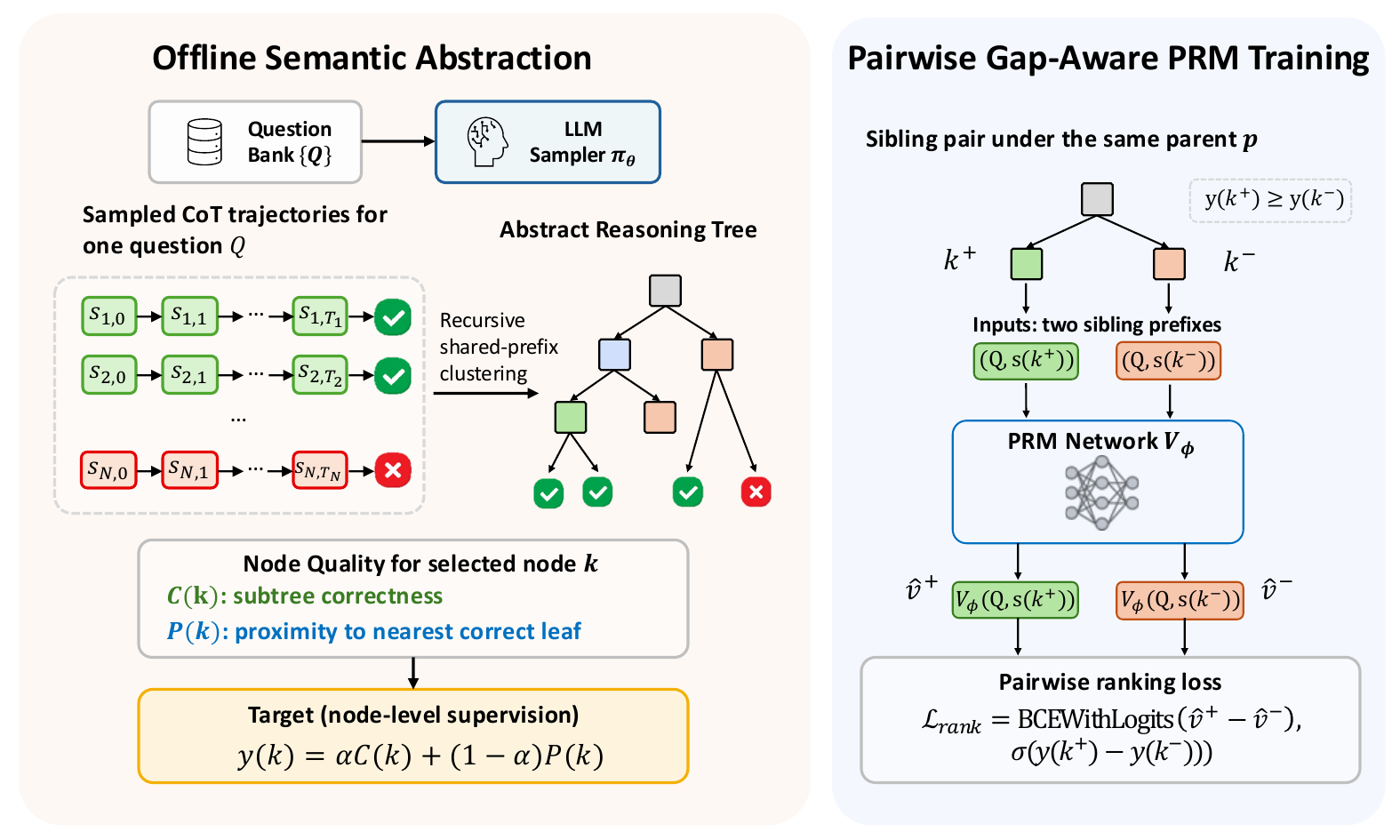}
    \caption{Overview of \our. Offline, we cluster same-depth reasoning steps into abstract states and train a value model on the resulting structure. At inference, the learned critic guides efficient exploration.}
    \label{fig:overview}
\end{figure}

Test-time inference scaling methods such as repeated sampling \cite{brown2024large} and self-consistency \cite{wang2022self} improve reliability by aggregating multiple reasoning chains, but their cost grows linearly with the number of sampled trajectories and they reuse little structure across solutions. Search-based approaches such as Tree-of-Thought (ToT) \cite{yao2023tree, yang2025qm} instead explore branching intermediate states, but they typically operate directly in free-form text space, where the number of possible continuations is large and many branches are superficially different but semantically redundant \cite{yao2022react}. This redundancy is especially pronounced in biomedical QA: recurring clinical reasoning operations, such as ruling out a diagnosis, identifying a mechanism, or eliminating an option, may appear across trajectories with diverse surface forms \cite{kartchner2023comprehensive}.

We propose \our, an offline-to-online framework for efficient biomedical reasoning that transforms sampled CoT trajectories into structured process supervision. Motivated by the semantic redundancy of medical reasoning traces, we study two research questions: (1) whether semantic abstraction of sampled reasoning trajectories can produce a compact and informative structure for process supervision; and (2) which node-level supervision signals best train a process critic for guiding test-time search. 

To address these questions, we first collect offline reasoning trajectories for each question and recursively cluster shared-prefix steps by semantic similarity using a sentence encoder. The resulting tree represents reasoning at the level of abstract clinical operations. We then derive node-level supervision from the tree structure by combining empirical subtree correctness with progress toward the nearest correct solution, and train a lightweight BERT-based process reward model (PRM) to score partial reasoning prefixes. At inference time, we use this learned critic to guide Monte Carlo Tree Search (MCTS) \cite{browne2012survey, jiang2025meds} through direct value evaluation, avoiding expensive terminal rollouts.


    
    

We evaluate \our on a comprehensive suite of biomedical QA benchmarks, comparing against single-pass CoT prompting, self-consistency, and alternative PRM-guided inference strategies. Across datasets, we find that semantic trajectory abstraction combined with value-guided inference improves reliability under fixed test-time budgets. We summarize our contributions as follows:

\begin{itemize}
    \item We show that biomedical chain-of-thought trajectories exhibit substantial semantic redundancy, and introduce a recursive depth-wise clustering framework that compresses diverse reasoning paths into a compact tree of abstract clinical reasoning states.
    
    \item We derive structured node-level supervision from the clustered reasoning tree, and demonstrate that progress toward the nearest correct solution provides a stronger and more stable training signal than subtree correctness.
    
    
    \item With a reward model 50 times smaller, we in some settings achieve comparable gain relative to self-consistency as larger state-of-the-art models, emerging as a more practical solution for resource-constrained clinical settings.
\end{itemize}

\section{Related Works}

\paragraph{Tree of Thought}
LLM reasoning has progressed from single chain-of-thought (CoT) traces \cite{wei2022chain} to test-time sampling and aggregation such as self-consistency \cite{wang2022self} and repeated sampling for inference scaling \cite{brown2024large}. Tree-of-Thought (ToT) reframes reasoning as explicit search over intermediate “thought” states, exploring and scoring alternative continuations in a tree \cite{yao2023tree}; Graph of Thoughts extends this to graph-structured reuse and recombination of states \cite{besta2024graph}. In medical settings, ToT-style search has been paired with multi-agent generation \cite{peng2025tree}, LLM-based judging to improve robustness \cite{yang2025qm} and concept bottleneck design for clinical trustworthiness \cite{nnamdi2026tree}.

\paragraph{Process Reward Models}
Process reward models (PRMs) supervise intermediate steps and can outperform outcome-only reward models when errors emerge late \cite{lightman2023let}. Supervision ranges from step-level binary labels \cite{zhang2025lessons} to richer preference or ranking objectives over partial solutions \cite{he2024advancing, zhang2024rest, guan2025rstar}. PRMs can guide inference and also serve as rewards for self-training or reinforcement learning to improve reasoning \cite{wang2024math}. In biomedicine, PRMs have been combined with retrieval and LLM-as-a-judge signals by conditioning scoring on retrieved evidence \cite{yun2025med}.

\paragraph{Value-guided Search for LLMs}
MCTS further improves LLM reasoning by searching a branching space of intermediate states and propagating feedback to guide expansion. Prior work uses PRM-guided tree search to collect higher-quality trajectories \cite{zhang2024rest} and applies MCTS for planning-style reasoning \cite{hao2023reasoning}, including settings that ensemble proposal models during expansion \cite{park2025ensembling}. AlphaZero-like methods replace costly rollouts with learned value estimates for selection and backup, improving search efficiency for decoding and training \cite{feng2023alphazero}. Combining test-time search with learned critics can yield substantial gains, including for smaller models \cite{guan2025rstar, jiang2025meds}.
\section{\our}
\subsection{Problem Formulation}
Given a question $Q$, we assume access to $N$ offline reasoning trajectories sampled from an LLM:
\begin{equation}
\tau_i = \big(s_{i,0}, s_{i,1}, \dots, s_{i,L_i-1}\big), \quad i=1,\dots,N,
\end{equation}
where $s_{i,t}$ denotes the text of the $t$-th reasoning step and $L_i$ is the number of steps in trajectory $i$.
Each trajectory is associated with an outcome label $y_i \in \{0,1\}$ indicating whether the final answer is correct.
Our goal is to transform these free-form trajectories into a compact reasoning tree, where nodes represent semantically equivalent intermediate steps at the same depth. This tree then provides structured process-level supervision for a lightweight critic used during test-time search.

\subsection{Clustered ToT Construction}

\begin{figure*}
    \centering
    \includegraphics[width=\linewidth]{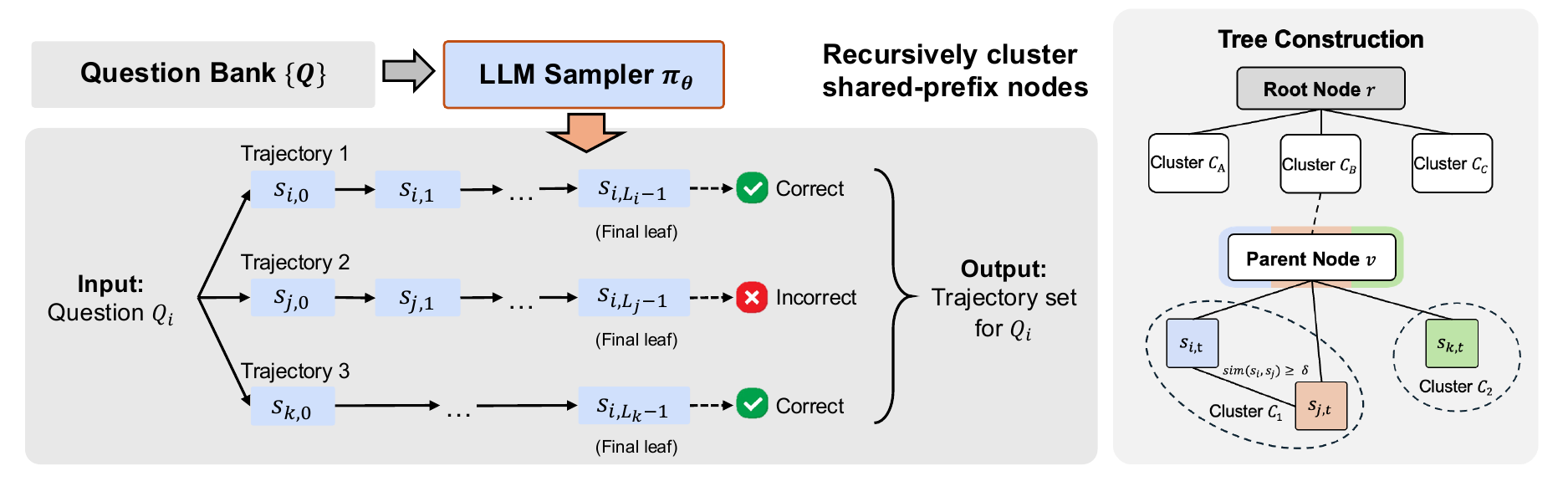}
    \caption{Iterative construction of tree nodes at the same depth under a shared parent. Steps are grouped when their embedding cosine similarity exceeds a threshold $\delta_{\mathrm{sim}}$.}
    \label{fig:cluster_constr}
\end{figure*}

We embed each reasoning step using a sentence encoder $f(\cdot)$:
\begin{equation}
\mathbf{h}_{i,t} = f(s_{i,t}).
\end{equation}
Step similarity is measured by cosine similarity between normalized embeddings. Given a threshold $\delta_{\mathrm{sim}}\in[0,1]$, two sibling steps are considered semantically equivalent if their similarity exceeds $\delta_{\mathrm{sim}}$.

The tree is constructed recursively in a prefix-preserving manner. At depth $t$, we only cluster steps from trajectories that share the same prefix-cluster assignments up to depth $t-1$. For a node $k$ at depth $t-1$, we collect the unique set of step strings $\mathcal{S}_t(k)$ from its valid immediate continuations. We then build an undirected threshold graph $G_t(k)$ over $\mathcal{S}_t(k)$, where an edge connects two step strings if their embedding cosine similarity exceeds $\delta_{\mathrm{sim}}$. The connected components of $G_t(k)$ define step clusters. Each cluster $C_j$ induces a child node $k_j$ at depth $t$, whose representative step is chosen as the most frequent step string in $C_j$.

This procedure is applied independently to every node at each depth. The recursion stops when no valid continuation remains or the maximum depth is reached. The resulting tree compactly represents the offline trajectory set: each root-to-leaf path corresponds to a sequence of abstract reasoning states, and each trajectory is assigned to exactly one node at every depth where it contains a step.

\subsection{Process Reward Model}

We assign each node $k$ in the reasoning tree two attributes that summarize the quality of continuing from its prefix. First, let $\mathbb{C}(k)\in[0,1]$ denote the fraction of correct leaf nodes in the subtree rooted at $k$, corresponding to the empirical success rate of the prefix. Second, let $\mathbb{P}(k)$ denote proximity to a correct answer, defined as the ratio between the depth of $k$ and the minimum depth among correct leaves in $k$'s subtree. These two attributes capture complementary notions of node quality: $\mathbb{C}(k)$ measures how promising the continuation space is, while $\mathbb{P}(k)$ measures how close the current prefix is to the nearest correct solution. In cases where there is no child correct leaf, $\mathbb{P}(k)$ is set to 0.

The node score is defined as
\begin{equation}
\label{eq:score_weight}
y(k) \;=\; \alpha\,\mathbb{C}(k) \;+\; (1-\alpha)\,\mathbb{P}(k),
\end{equation}

where $\alpha\in[0,1]$ controls the trade-off between empirical success rate and progress toward the nearest correct leaf.

We construct training pairs from sibling prefixes that share the same parent. For a parent node $p$ with children $k_i$ and $k_j$, we compute their scalar scores $y(k_i)$ and $y(k_j)$. Each pair is ordered so that $y(k^+)\ge y(k^-)$, yielding a positive prefix $k^+$ and a negative prefix $k^-$. We then train a critic $V_\phi$. Given question $Q$ and prefix $s(k)$ for node $k$, the PRM outputs
\begin{equation}
\hat{v}(k) = V_\phi(Q,s(k))
\end{equation}

Let $\Delta_{+-}=y(k^+)-y(k^-)$, $r_\phi=\hat{v}(k^+)-\hat{v}(k^-)$, and $\eta_{+-}=\sigma(\Delta_{+-})$. The PRM is trained to prefer the higher-scoring sibling using a gap-aware ranking loss:
\begin{equation}
\label{eq:loss}
\mathcal{L}_{\mathrm{rank}}
=
\operatorname{BCEWithLogits}\!\left(r_\phi, \eta_{+-}\right).
\end{equation}


This pairwise objective matches the downstream use of the PRM, where candidate continuations are compared under the same parent prefix.

\subsection{Process-Critic MCTS}
\label{sec:demcts}

\begin{figure*}
    \centering
    \includegraphics[width=0.9\linewidth]{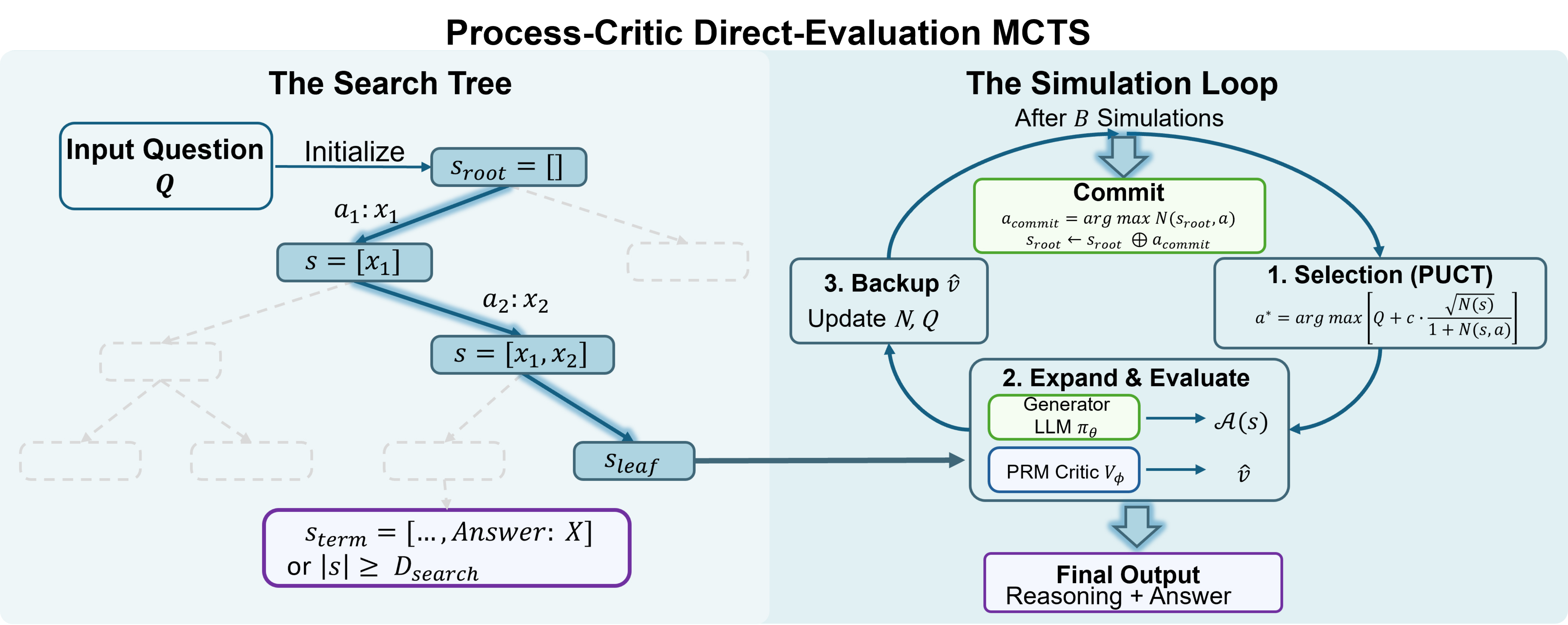}
    \caption{Test-time process-critic MCTS for reasoning. New leaf nodes are expanded and evaluated, and the value is backed up through the tree.}
    \label{fig:mcts}
\end{figure*}

At test time, we construct a search tree on the fly, where each node is a partial reasoning prefix and its children are candidate next steps generated by the LLM backbone. Figure~\ref{fig:mcts} illustrates the procedure. Children are generated once and cached for reuse. Unlike rollout-based MCTS, we do not simulate each path to a terminal answer; instead, the learned critic directly evaluates newly expanded leaves.

Given a question $Q$, a state is a reasoning prefix $s=[x_1,\dots,x_t]$, and an action $a\in\mathcal{A}(s)$ is a candidate next step. A state is terminal if it emits an explicit answer or reaches the depth limit. When $s$ is first expanded, we query the backbone $\pi_\theta$ to propose $K$ candidate next steps, remove duplicate or empty candidates, and cache the remaining set as $\mathcal{A}(s)$. When a simulation reaches an unexpanded or terminal node, the critic $V_\phi$ assigns its value.

The leaf value is backed up along the selected path to update visit counts $N$ and action values $Q$. After $B$ simulations, we commit the root action with the largest visit count and repeat the search from the new root. If no explicit final answer is produced, we apply an answer-forcing prompt. Additional details are listed in Appendix~\ref{appendix:mcts}.

\section{Experiments}
\subsection{Experimental Setups}

Our experiments evaluate whether a clustering-derived PRM can effectively guide step-wise reasoning for medical question answering. We study two PRM-guided inference strategies: Best-of-N and direct-evaluation MCTS, and compare them against standard CoT and state-of-the-art medical QA reward model baselines. To assess both in-domain effectiveness and cross-model robustness, we study backbones that are used to collect the offline training trajectories and backbones that do not. We further evaluate on four in-distribution (ID) datasets and three out-of-distribution OOD datasets to characterize generalization across both models and benchmarks.

\paragraph{Datasets.}
We construct the offline trajectory corpus from four medical QA benchmarks: PubMedQA~\cite{jin2019pubmedqa}, MedQA~\cite{jin2021disease}, MedMCQA~\cite{pal2022medmcqa}, and BioASQ~\cite{krithara2023bioasq}. PubMedQA, MedQA, and MedMCQA are multiple-choice question (MCQ) datasets, whereas BioASQ contains open-ended biomedical questions. The dataset splits follow Table~\ref{tab:data_splits}, yielding approximately 103k questions for training and 2.4k for validation.

Evaluation is conducted on the ID test sets of these four datasets, comprising 3.8k questions in total, and on three OOD benchmarks: MedQuAD~\cite{BenAbacha-BMC-2019}, MedXpertQA-Text~\cite{zuo2025medxpertqa}, and MMLU-Med~\cite{hendryckstest2021}, comprising 5.2k questions in total.

\paragraph{Backbones.}
We use three mid-sized generalist models for offline trajectory collection: Qwen3-8B, Llama-3.1-8B, and Gemma-2-9B. To evaluate cross-backbone robustness, we additionally test two backbones, OLMo-3-7B and Phi-3.5, whose trajectories are not used during PRM training. The latter two are smaller, providing a stricter test of whether the learned process critic can transfer to less capable or differently trained generators.
\paragraph{Baselines.}
We compare \our against the following baselines:

$\diamond$~\textbf{Self-Consistency} \cite{wang2022self}: We sample multiple solutions and aggregate them using majority voting for MCQs and score averaging for open-ended QA.

$\diamond$~\textbf{Best-of-N} \cite{nakano2021webgpt}: We independently sample $N$ complete reasoning trajectories for each question. Our reward model $V_\phi$ scores each trajectory, and the solution with the highest score is selected as the final answer.

$\diamond$~\textbf{Med-PRM} \cite{yun2025med}: We use the publicly released Med-PRM reward model to score the same $N$ trajectories generated by each backbone and select the highest-scoring solution. We additionally report results using its original Meerkat-8B policy model.

$\diamond$~\textbf{MedS3} \cite{jiang2025meds}: We use the publicly released MedS3 reward model to score the same $N$ trajectories generated by each backbone and select the highest-scoring solution. We also report results using its original policy model.

\paragraph{Evaluation metrics.}
We evaluate MCQ performance using accuracy and open-ended QA performance using BERTScore F1 and ROUGE-L F1.

\subsection{Main Results}

Table~\ref{tab:main_results} reports performance under a fixed inference budget of 4,096 tokens. Results are averaged over five random seeds and presented with standard deviations. Detailed per-dataset results are provided in Tables~\ref{tab:detail_res_id} and~\ref{tab:detail_res_ood} in the Appendix. The aggregate scores in Table~\ref{tab:main_results} are weighted by the number of evaluation samples in each dataset.

\begin{table*}[t]
\centering
\footnotesize
\setlength{\tabcolsep}{3pt}
\providecommand{\stdpm}[1]{\,{\tiny$\pm$#1}}
\renewcommand{\arraystretch}{0.9}
\providecommand{\ours}{\cellcolor{gray!12}}
\begin{tabular*}{\textwidth}{@{\extracolsep{\fill}}lllcccccc@{}}
\toprule
\multirow{3}{*}{Backbone} & \multirow{3}{*}{Reward} & \multirow{3}{*}{Search}
& \multicolumn{3}{c}{In-distribution} & \multicolumn{3}{c}{Out-of-distribution} \\
\cmidrule(lr){4-6}\cmidrule(lr){7-9}
& & & MCQ & \multicolumn{2}{c}{Open-ended} & MCQ & \multicolumn{2}{c}{Open-ended} \\
\cmidrule(lr){5-6}\cmidrule(lr){8-9}
& & & Acc $\uparrow$ & BERTSc. $\uparrow$ & RougeL $\uparrow$ & Acc $\uparrow$ & BERTSc. $\uparrow$ & RougeL $\uparrow$ \\
\midrule
\multirow{7}{*}{Qwen3-8B}
& --                                       & Self-Cons.       & 54.9\stdpm{0.8} & 86.5\stdpm{1.3} & 12.8\stdpm{0.5} & 53.7\stdpm{0.9} & \textbf{87.0}\stdpm{1.4} & \underline{12.5}\stdpm{0.2} \\
&                                          & Best-of-N        & \textbf{59.8}\stdpm{1.9} & 83.6\stdpm{1.2} & 17.2\stdpm{0.6} & \textbf{62.3}\stdpm{1.1} & 70.2\stdpm{4.4} & 11.1\stdpm{0.8} \\
& \multirow{-2}{*}{MedS3}                  & MCTS             & 56.7\stdpm{1.7} & 88.2\stdpm{1.0} & \textbf{27.8}\stdpm{2.3} & 54.2\stdpm{1.1} & 84.3\stdpm{0.1} & 6.6\stdpm{0.2} \\
&                                          & Best-of-N        & 56.8\stdpm{1.8} & 83.7\stdpm{1.5} & 16.4\stdpm{1.0} & 59.8\stdpm{1.8} & 80.6\stdpm{2.4} & \textbf{12.7}\stdpm{0.9} \\
& \multirow{-2}{*}{Med-PRM}                & MCTS             & 55.4\stdpm{2.0} & \textbf{89.1}\stdpm{0.5} & \underline{27.7}\stdpm{3.1} & 53.4\stdpm{0.7} & \underline{84.4}\stdpm{0.1} & 6.3\stdpm{0.2} \\
& \ours                                    & \ours Best-of-N  & \ours \underline{57.7}\stdpm{4.3} & \ours 85.5\stdpm{2.3} & \ours 16.2\stdpm{0.3} & \ours 60.1\stdpm{2.2} & \ours 82.7\stdpm{0.6} & \ours 12.3\stdpm{0.3} \\
& \ours \multirow{-2}{*}{\textbf{MedCoast}} & \ours MCTS       & \ours 57.5\stdpm{2.3} & \ours \underline{88.9}\stdpm{0.2} & \ours 26.7\stdpm{1.5} & \ours \underline{60.8}\stdpm{1.3} & \ours 84.2\stdpm{0.8} & \ours 7.4\stdpm{0.1} \\
\midrule
\multirow{7}{*}{Llama-3.1-8B}
& --                                       & Self-Cons.       & \underline{57.3}\stdpm{1.3} & 75.0\stdpm{0.8} & 13.7\stdpm{0.2} & 50.1\stdpm{1.5} & 83.0\stdpm{0.1} & 14.6\stdpm{0.2} \\
&                                          & Best-of-N        & \textbf{61.1}\stdpm{1.2} & 84.0\stdpm{1.8} & 11.3\stdpm{0.8} & \textbf{60.3}\stdpm{0.8} & \underline{86.0}\stdpm{1.9} & \underline{15.9}\stdpm{0.2} \\
& \multirow{-2}{*}{MedS3}                  & MCTS             & 47.2\stdpm{3.2} & 86.9\stdpm{0.5} & \underline{18.0}\stdpm{2.9} & 51.1\stdpm{1.8} & 82.3\stdpm{0.2} & 4.8\stdpm{0.1} \\
&                                          & Best-of-N        & \underline{57.3}\stdpm{2.6} & 84.4\stdpm{0.9} & 10.9\stdpm{0.8} & \underline{55.8}\stdpm{2.4} & \textbf{87.2}\stdpm{0.1} & \textbf{16.4}\stdpm{0.5} \\
& \multirow{-2}{*}{Med-PRM}                & MCTS             & 46.4\stdpm{3.1} & \textbf{87.3}\stdpm{0.1} & 17.9\stdpm{0.0} & 52.1\stdpm{0.9} & 82.4\stdpm{0.5} & 4.4\stdpm{0.1} \\
& \ours                                    & \ours Best-of-N  & \ours 53.0\stdpm{2.9} & \ours 86.9\stdpm{0.4} & \ours 11.6\stdpm{1.2} & \ours 55.4\stdpm{1.4} & \ours \underline{86.0}\stdpm{1.0} & \ours 15.7\stdpm{0.2} \\
& \ours \multirow{-2}{*}{\textbf{MedCoast}} & \ours MCTS       & \ours 48.3\stdpm{3.9} & \ours \underline{87.2}\stdpm{0.5} & \ours \textbf{18.8}\stdpm{3.0} & \ours 53.6\stdpm{2.5} & \ours 82.2\stdpm{0.2} & \ours 4.5\stdpm{0.6} \\
\midrule
\multirow{7}{*}{Gemma-2-9B}
& --                                       & Self-Cons.       & 35.9\stdpm{2.9} & 75.9\stdpm{4.3} & 18.4\stdpm{0.7} & 37.4\stdpm{0.6} & 46.7\stdpm{1.3} & 6.4\stdpm{2.6} \\
&                                          & Best-of-N        & 46.8\stdpm{2.6} & 62.8\stdpm{3.0} & 14.8\stdpm{0.5} & \underline{55.7}\stdpm{0.9} & 59.5\stdpm{5.3} & 7.9\stdpm{0.7} \\
& \multirow{-2}{*}{MedS3}                  & MCTS             & \underline{47.5}\stdpm{1.6} & 85.4\stdpm{0.6} & 23.5\stdpm{0.9} & 49.2\stdpm{1.9} & \underline{81.4}\stdpm{0.1} & 3.6\stdpm{0.4} \\
&                                          & Best-of-N        & 45.0\stdpm{3.0} & 73.2\stdpm{2.5} & 13.7\stdpm{0.3} & 53.3\stdpm{0.9} & 70.4\stdpm{3.3} & \textbf{9.7}\stdpm{0.3} \\
& \multirow{-2}{*}{Med-PRM}                & MCTS             & \textbf{48.3}\stdpm{1.5} & \underline{86.8}\stdpm{0.4} & \textbf{25.8}\stdpm{1.4} & 49.5\stdpm{1.6} & \textbf{82.3}\stdpm{0.6} & 3.7\stdpm{0.2} \\
& \ours                                    & \ours Best-of-N  & \ours 44.5\stdpm{2.7} & \ours 74.4\stdpm{0.8} & \ours 16.9\stdpm{0.7} & \ours 54.8\stdpm{2.8} & \ours 73.9\stdpm{4.7} & \ours \underline{9.3}\stdpm{0.4} \\
& \ours \multirow{-2}{*}{\textbf{MedCoast}} & \ours MCTS       & \ours \textbf{48.3}\stdpm{1.3} & \ours \textbf{87.3}\stdpm{1.2} & \ours \underline{24.8}\stdpm{1.9} & \ours \textbf{56.7}\stdpm{2.4} & \ours 80.8\stdpm{3.1} & \ours 3.5\stdpm{0.5} \\
\midrule
\multirow{7}{*}{OLMo-3-7B}
& --                                       & Self-Cons.       & \textbf{41.6}\stdpm{2.1} & 22.1\stdpm{3.1} & 6.2\stdpm{1.0} & \textbf{45.5}\stdpm{2.0} & 33.0\stdpm{2.9} & \underline{5.0}\stdpm{0.8} \\
&                                          & Best-of-N        & 37.1\stdpm{0.8} & 43.0\stdpm{3.5} & 8.2\stdpm{0.5} & 35.8\stdpm{1.4} & 15.4\stdpm{2.5} & 2.3\stdpm{0.6} \\
& \multirow{-2}{*}{MedS3}                  & MCTS             & \underline{40.9}\stdpm{1.7} & \underline{84.8}\stdpm{1.7} & 15.7\stdpm{1.2} & 43.8\stdpm{1.6} & 79.3\stdpm{2.4} & 4.0\stdpm{0.4} \\
&                                          & Best-of-N        & 31.4\stdpm{4.0} & 42.7\stdpm{3.5} & 7.0\stdpm{0.4} & 34.3\stdpm{3.6} & 30.5\stdpm{4.6} & 4.0\stdpm{1.0} \\
& \multirow{-2}{*}{Med-PRM}                & MCTS             & 39.4\stdpm{1.6} & 84.4\stdpm{1.5} & \textbf{20.3}\stdpm{1.2} & \underline{44.7}\stdpm{0.6} & \underline{80.6}\stdpm{2.0} & 4.5\stdpm{0.6} \\
& \ours                                    & \ours Best-of-N  & \ours 34.9\stdpm{5.0} & \ours 80.8\stdpm{1.9} & \ours 11.9\stdpm{1.7} & \ours 33.4\stdpm{3.7} & \ours 73.2\stdpm{5.4} & \ours \textbf{11.6}\stdpm{0.8} \\
& \ours \multirow{-2}{*}{\textbf{MedCoast}} & \ours MCTS       & \ours 40.5\stdpm{2.8} & \ours \textbf{85.3}\stdpm{1.7} & \ours \underline{17.1}\stdpm{0.8} & \ours 44.5\stdpm{2.8} & \ours \textbf{81.8}\stdpm{0.7} & \ours 4.8\stdpm{0.4} \\
\midrule
\multirow{7}{*}{Phi-3.5-mini}
& --                                       & Self-Cons.       & 40.4\stdpm{3.6} & 27.7\stdpm{1.9} & 1.0\stdpm{0.2} & 40.5\stdpm{1.4} & 51.0\stdpm{2.8} & 5.4\stdpm{0.4} \\
&                                          & Best-of-N        & 42.5\stdpm{0.9} & 42.7\stdpm{0.7} & 4.3\stdpm{0.1} & 37.0\stdpm{3.0} & 37.3\stdpm{2.0} & 3.9\stdpm{0.4} \\
& \multirow{-2}{*}{MedS3}                  & MCTS             & \underline{47.1}\stdpm{2.4} & 87.1\stdpm{1.1} & \textbf{22.2}\stdpm{1.7} & \underline{47.8}\stdpm{3.2} & \underline{79.0}\stdpm{2.8} & 4.0\stdpm{0.3} \\
&                                          & Best-of-N        & 46.8\stdpm{2.8} & 81.6\stdpm{2.2} & 5.5\stdpm{0.8} & \textbf{49.9}\stdpm{1.7} & 76.2\stdpm{2.0} & \textbf{8.2}\stdpm{0.8} \\
& \multirow{-2}{*}{Med-PRM}                & MCTS             & 45.9\stdpm{1.1} & \underline{87.5}\stdpm{0.8} & \underline{20.6}\stdpm{3.0} & 46.5\stdpm{2.6} & \textbf{80.6}\stdpm{0.5} & 3.7\stdpm{0.5} \\
& \ours                                    & \ours Best-of-N  & \ours 41.7\stdpm{1.9} & \ours 76.6\stdpm{2.2} & \ours 5.3\stdpm{0.5} & \ours 41.8\stdpm{2.2} & \ours 73.1\stdpm{3.8} & \ours \underline{6.9}\stdpm{1.0} \\
& \ours \multirow{-2}{*}{\textbf{MedCoast}} & \ours MCTS       & \ours \textbf{47.3}\stdpm{2.2} & \ours \textbf{87.9}\stdpm{0.3} & \ours 19.3\stdpm{1.5} & \ours 46.1\stdpm{1.5} & \ours 78.9\stdpm{1.8} & \ours 4.0\stdpm{0.4} \\
\midrule
MedS3 Policy   & MedS3   & Best-of-N & 60.4\stdpm{3.1} & 76.2\stdpm{0.2} & 16.2\stdpm{0.6} & 57.4\stdpm{1.4} & 73.9\stdpm{2.0} & 12.8\stdpm{0.8} \\
Med-PRM Policy & Med-PRM & Best-of-N & 59.8\stdpm{0.6} & 85.1\stdpm{0.9} & 10.5\stdpm{0.3} & 56.7\stdpm{2.0} & 73.5\stdpm{1.8} & 9.1\stdpm{0.5} \\
\bottomrule
\end{tabular*}
\caption{Performance on in-distribution and out-of-distribution medical QA benchmarks under a 4096-token inference budget. MCQ columns are count-weighted accuracies; open-ended columns report BioASQ (ID) and MedQuAD (OOD). Within each backbone, \textbf{bold} marks the best and \underline{underline} the second-best result per column.}
\label{tab:main_results}
\end{table*}

\paragraph{In-distribution performance.}
Across the ID medical benchmarks, the cluster-based PRM generally improves open-source generalist backbones over self-consistency. MCQ performance is more mixed: results decrease on Llama-3.1-8B, while Qwen3-8B and Gemma-2-9B achieve gains of up to 18 points. Compared to state-of-the-art reward-guided approaches, \our ranks first or second in most settings. For open-ended BioASQ performance, our process critic consistently improves across all three backbones.

\paragraph{Out-of-distribution performance.}
On the OOD benchmarks, Qwen3-8B, Llama-3.1-8B, and Gemma-2-9B obtain smaller but generally positive gains from PRM-guided inference. Notably, on MedXpertQA, \our approximately doubles the self-consistency accuracy of Gemma-2-9B. In many of these settings, MedS3 and Med-PRM achieve comparable performance.

Smaller backbones, including OLMo-3 and Phi-3.5, benefit substantially from \our across the seven benchmarks. In contrast, MedS3 and Med-PRM do not provide improvements of a similar magnitude for these two backbones, allowing \our to outperform them by wider margins. Across all five backbones, \our exhibits more stable performance, suggesting stronger cross-backbone generalizability. For MCQ accuracy, the standard deviation across backbones is 4.47 for \our, compared with 9.33 for MedS3 and 8.40 for Med-PRM. The difference is even larger for open-ended BERTScore, for which the corresponding standard deviations are 1.65, 24.40, and 19.76, respectively.

\paragraph{Comparison to state-of-the-art.}

When paired with their respective policy models, the MedS3 and Med-PRM reward models achieve strong performance, particularly on in-distribution MCQ benchmarks. Nevertheless, \our offers two practical advantages. First, as discussed above, both existing reward models are more sensitive to the choice of backbone, whereas \our demonstrates more stable cross-backbone performance. Second, both reward models contain approximately 8B parameters. Deploying them alongside an 8B policy model therefore imposes substantial memory and hardware requirements. In contrast, our BERT-based PRM contains only 149M parameters, making \our a more accessible solution for resource-constrained clinical settings.



\subsection{Ablation Studies}

\paragraph{Semantic clustering.}

Table~\ref{tab:agg_cluster_ablation} summarizes the effect of semantic clustering, reporting accuracy for MCQ benchmarks and BERTScore for open-ended benchmarks. Detailed per-dataset results are provided in Table~\ref{tab:cluster_compare_full}. All results are averaged over five random seeds. The clustered variants generally outperform their non-clustered counterparts, revealing that semantic clustering contributes to the performance gains of \our.

\begin{table}[t]
\centering
\small
\setlength{\tabcolsep}{2.5pt}
\renewcommand{\arraystretch}{1.05}
\begin{tabular}{llcccc}
\toprule
\multirow{2}{*}{Inf. Method} & \multirow{2}{*}{Cluster}
& \multicolumn{2}{c}{In-distribution} & \multicolumn{2}{c}{Out-of-distribution} \\
\cmidrule(lr){3-4}\cmidrule(lr){5-6}
& & MCQ & BioASQ & MCQ & MedQuAD \\
\midrule
\multirow{3}{*}{Best-of-N}
& Cluster     & 45.7 & 80.8 & 46.4 & 77.8 \\
& Uncluster   & 44.6 & 80.5 & 47.8 & 72.4 \\
\cmidrule(lr){2-6}
& $\Delta$      & +1.1 & +0.4 & -1.4 & +5.3 \\
\cmidrule(lr){1-6}
\multirow{3}{*}{MCTS}
& Cluster     & 49.1 & 87.3 & 49.6 & 81.6 \\
& Uncluster   & 47.9 & 86.6 & 48.5 & 82.0 \\
\cmidrule(lr){2-6}
& $\Delta$      & +1.2 & +0.7 & +1.2 & -0.4 \\
\bottomrule
\end{tabular}
\caption{Cluster vs.\ uncluster ablation for Best-of-N and MCTS, aggregated over datasets and averaged across the five backbones. Numbers are averages across five random seeds.}
\label{tab:agg_cluster_ablation}
\end{table}

\paragraph{Node scoring.}

We ablate two key design choices in constructing PRM supervision: the semantic-similarity threshold $\delta_{\mathrm{sim}}$ used to cluster individual trajectories into the structured reasoning tree, and the node-level target used to supervise the critic. For the node target, we compare three settings: proximity-only supervision, which uses progress toward the nearest correct answer; correctness-only supervision, which uses the empirical correctness ratio of the subtree; and mixed supervision, which averages progress and subtree correctness. These correspond to subtree correctness $\alpha$ of $0$, $1.0$, and $0.5$, respectively. For the semantic clustering, we evaluate thresholds $\delta_{\mathrm{sim}}$ at $0.75$, $0.80$, and $0.85$.

\begin{figure}[t]
    \centering
    \includegraphics[width=\linewidth]{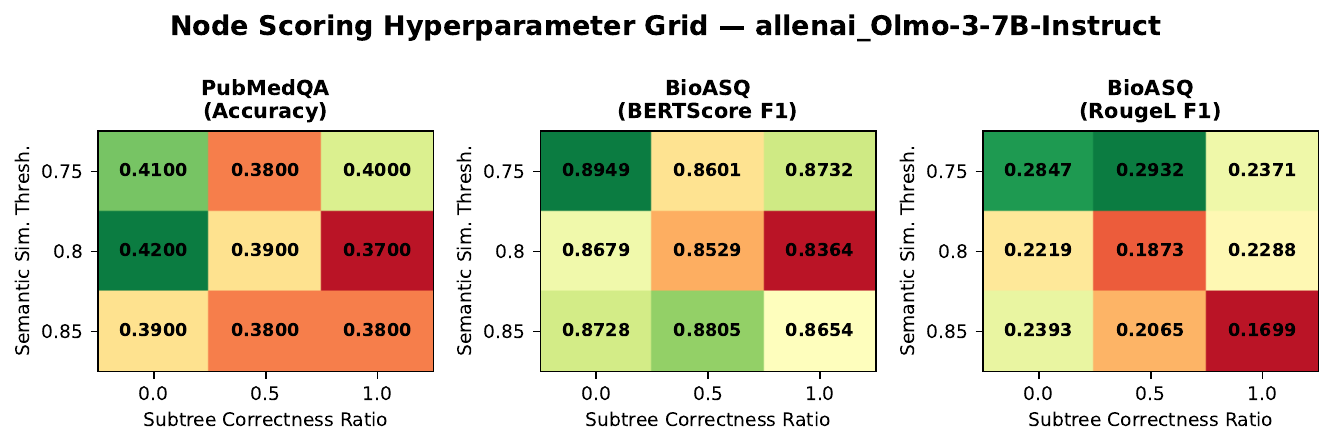}
    \caption{Grid-view ablation of node scoring design on PubMedQA and BioASQ using OLMo-3-7B-Instruct with direct-evaluation MCTS ($B=6$ and $K=4$).}
    \label{fig:grid}
\end{figure}

Figure~\ref{fig:grid} reports the resulting performance grid for direct-evaluation MCTS with the out-of-distribution OLMo-3 backbone. Two trends emerge. First, the choice of node-level supervision has a substantial effect on performance. In particular, incorporating progress toward a correct answer is more beneficial than relying only on subtree correctness, suggesting that dense progress signals provide more informative supervision for intermediate reasoning states. Second, performance is comparatively less sensitive to the clustering threshold, which indicates that the method is robust to moderate changes in the semantic granularity of the offline reasoning tree. Appendix \ref{appendix:inf_sensitivity} presents additional results across all seven datasets with Qwen3 and Phi-3.5 backbones. Statistical analysis reveals that the effect size of $\alpha$ is indeed higher than $\delta_{\mathrm{sim}}$.

\paragraph{Critic training objective.}

\begin{figure}[t]
    \centering
    \includegraphics[width=\linewidth]{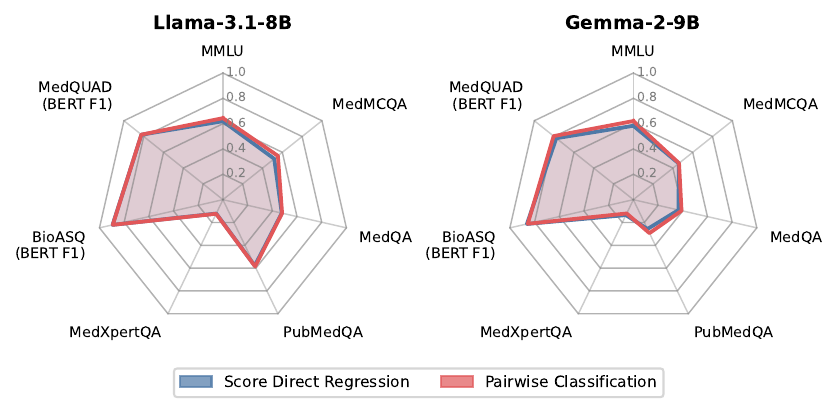}
    \caption{Radar plots comparing two models across seven datasets, showing that pairwise gap classification slightly improves downstream PRM-guided MCTS inference.}
    \label{fig:gap_loss}
\end{figure}

In \our, we construct positive--negative sibling pairs and pass their corresponding reasoning prefixes separately through the BERT-based critic. The objective in Equation~\ref{eq:loss} is formulated as a pairwise classification problem: within each pair, the child with the higher ground-truth node score is treated as the preferred continuation. A more naive alternative is to train the process critic by directly regressing each node's absolute score. Our experiments suggest that this shift from direct score regression is important. Figure~\ref{fig:gap_loss} shows that direct regression to individual child-node scores slightly degrades downstream performance. We attribute this to noise in the absolute node-level targets, which can obscure the relative patterns that distinguish promising reasoning prefixes from weaker ones.

By contrast, the pairwise gap objective more closely matches the inference-time setting, where newly expanded candidate steps are compared against one another. Optimizing this objective encourages the critic to preserve the correct relative ranking among sibling nodes, which is the behavior most relevant for greedy selection, beam search, and MCTS-based exploration. In Figure~\ref{fig:gap_loss}, the supervision setting is fixed to a clustering cutoff of $0.8$ and a subtree-correctness weight of $0$, corresponding to progress-only node scoring.

\subsection{Scaling Analysis}

\begin{figure}[t]
    \centering
    \includegraphics[width=\linewidth]{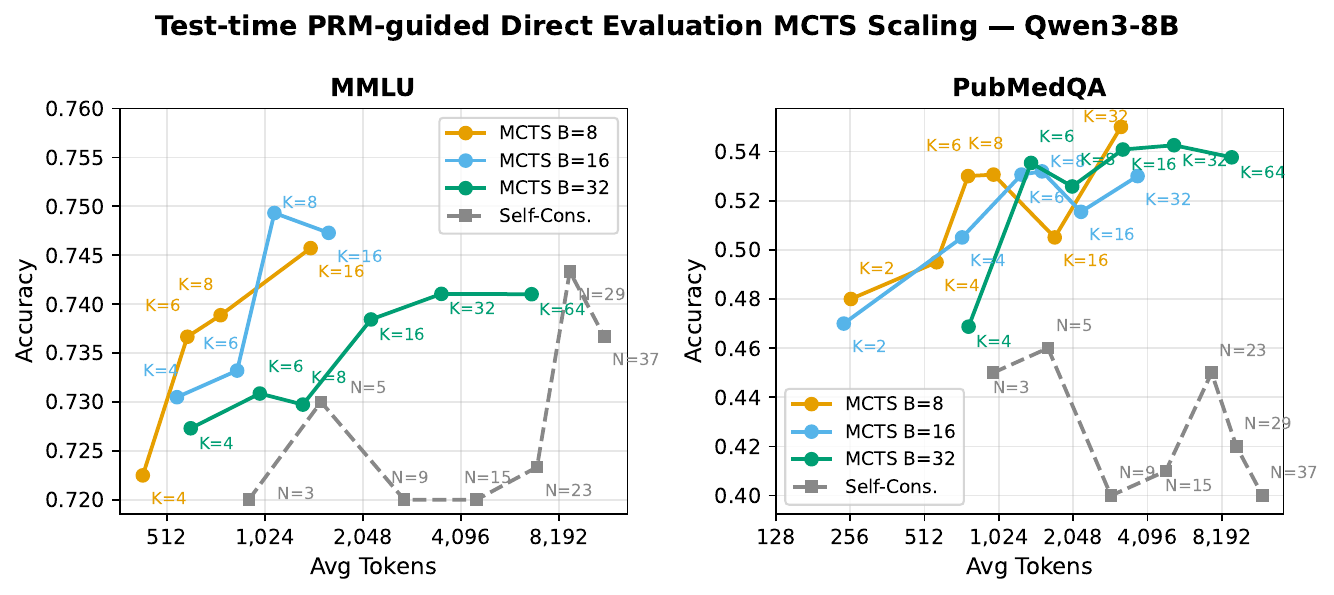}
    \caption{Accuracy as a function of token budgets. Varying the MCTS simulations ($B$) and candidate branching ($K$) increases the total number of generated tokens and improves accuracy.}
    \label{fig:scale_mcts}
\end{figure}

Although test-time MCTS is not the primary novelty of this work, we analyze its scaling behavior to evaluate whether the learned PRM provides a robust guidance signal under increasing inference budgets. Figure~\ref{fig:scale_mcts} shows how test-time compute affects MCTS performance with Qwen3-8B on the MMLU-Med and PubMedQA datasets. We vary two inference-time hyperparameters: the number of MCTS simulations ($B$) and the number of candidate next steps generated per node ($K$). Increasing either parameter increases the total number of generated tokens, allowing us to measure accuracy as a function of test-time compute while keeping the backbone model fixed. The dashed line denotes the self-consistency baseline evaluated with multiple sampled completions.

\our outperforms self-consistency across a range of token budgets. Moreover, self-consistency exhibits weaker scaling behavior, because its gains diminish quickly and tend to plateau at substantially lower token counts. In contrast, PRM-guided search continues to benefit from additional inference compute. We show that the learned critic provides a useful signal for allocating computation toward more promising reasoning branches rather than simply sampling more independent trajectories.

\subsection{Qualitative Evaluation}

We perform a qualitative error analysis to characterize failure modes of process-critic MCTS. We randomly sample 50 inference trajectories and manually inspect both the generated reasoning chains and final outputs. This analysis identifies four recurring failure patterns:

$\diamond$~Repetitive reasoning steps: the model emits a step $s_t$ that duplicates an earlier generated step $s_i$ for some $i < t$.

$\diamond$~Medical knowledge gap: the model lacks the domain knowledge needed to support the correct answer, leading to clinically incorrect or misaligned reasoning.

$\diamond$~Reasoning trace error: the reasoning trace contains medically relevant evidence and may implicitly support the correct answer, but the model fails to select the optimal final answer.

$\diamond$~Missing answer emission: the model produces a coherent multi-step reasoning trace but fails to emit a final answer.

Details of the human evaluation can be found in Appendix \ref{appendix:human_eval_guide}. Table~\ref{tab:human_analysis} reports the distribution of these failure patterns for correctly and incorrectly answered questions. Among incorrect inferences, medical knowledge gaps are the dominant failure mode, indicating that process critic can guide the model toward more promising reasoning trajectories but cannot substitute for domain-specific knowledge acquisition. This limitation is also reflected in cases where knowledge errors occur despite the final answer being correct. Repetition is another common failure mode across both correct and incorrect inferences. Representative examples for each failure category are provided in the Appendix \ref{appendix:wrong}.

In particular, Table~\ref{tab:fail_rep} illustrates cases with repeated reasoning steps. Further analysis of the same two questions under one-pass CoT prompting with the same backbone (Table~\ref{tab:rep_q_cot}) suggests that this looping behavior may reflect an inherent tendency of the base model, rather than an artifact introduced by \our inference procedure. More broadly, this failure mode suggests that medical reasoning may not always be well captured by a strictly linear Tree-of-Thought structure. Unlike mathematical problem solving, clinical reasoning often integrates evidence from multiple sources, and the decision process may be less ordinal. This observation points to Graph-of-Thoughts as a promising extension. Given our finding that proximity to a correct answer provides a more effective supervisory signal than subtree correctness alone, the same principle could be used to guide value-based exploration over graph-structured reasoning states.

\begin{table}[t]
\centering
\small
\setlength{\tabcolsep}{2.5pt}
\renewcommand{\arraystretch}{1.15}
\begin{tabular}{lcccc}
\toprule
\textbf{Output Group} & \textbf{Repeat} & \textbf{Know.} & \textbf{Trace Err.} & \textbf{Miss Ans.} \\
\midrule
Correct ($n=20$)   & 20.0\% & 10.0\% & 0.0\% & 0.0\% \\
Incorrect ($n=30$) & 43.3\% & 63.3\% & 6.7\% & 3.3\% \\
\bottomrule
\end{tabular}
\caption{Reasoning failure patterns among correct and incorrect model outputs. Percentages are computed within each output group, and one output may contain multiple issues.}
\label{tab:human_analysis}
\end{table}
\section{Conclusion}

We introduced \our, an offline-to-online framework that compresses sampled chain-of-thought trajectories into compact process supervision for test-time biomedical reasoning. By recursively clustering semantically similar reasoning steps under shared prefixes, deriving node-level supervision based on proximity to the nearest correct solution, and training a lightweight process reward model with a pairwise gap-aware objective, \our transforms noisy reasoning samples into a learning signal aligned with the sibling-level comparisons performed during inference. Across seven biomedical question-answering benchmarks, \our improves over self-consistency under fixed token budgets and, in several settings, achieves performance comparable to state-of-the-art reward models that are over 50 times larger.


\section*{Limitations}

Our method trades a substantial one-time offline construction cost for more efficient and effective test-time inference. Building the offline trajectory corpus, clustering reasoning steps, and training the PRM required approximately 300 GPU hours in our experiments. For BioASQ, we binarized the open-ended answers using ROUGE-L scores. We view this as a practical compromise that allows open-ended questions to be incorporated alongside MCQs, thereby broadening the training distribution for the PRM. Our experiments also focus primarily on reasoning over question text alone. In real clinical settings, reliable medical reasoning should be grounded in external evidence, including clinical guidelines, biomedical literature, and patient-specific records. Integrating retrieval into both the proposal model and the process critic is therefore an important direction for future work. In addition, all open-source models are evaluated under quantization to fit available GPU memory. Quantization may reduce generation quality, alter the distribution of candidate reasoning steps, and affect the search landscape explored by MCTS and beam search. As a result, the reported absolute performance may underestimate what could be achieved with full-precision models and larger compute budgets.

\section*{Ethics Statements}
This work is conducted entirely on publicly available benchmarks, including BioASQ, MedQA, MedMCQA, PubMedQA, MedXpertQA, MedQuAD, and MMLU. These datasets do not contain patient-identifying information, and our study does not collect, store, or process personal user data. We use the datasets only for aggregate benchmark evaluation, respect their licenses and access conditions, and do not redistribute restricted dataset content or individual examples beyond what is permitted.

Although several benchmarks are medical in nature, the proposed system is intended only for research on language-model reasoning and test-time inference. It is not validated for clinical deployment and must not be used to provide diagnosis, treatment recommendations, or medical decision support. Model outputs may contain factual errors, unsupported reasoning, or unsafe conclusions, especially under distribution shift. In addition, benchmark questions are imperfect proxies for real-world medical practice. They may reflect exam-oriented reasoning, geographic and specialty biases, outdated clinical assumptions, or limited representation of diverse patient populations. Therefore, improvements on these benchmarks should be interpreted as evidence of better benchmark reasoning performance, not clinical reliability or readiness for patient care.



\bibliography{custom}

\clearpage
\appendix

\newpage
\section{Appendix}
\label{sec:appendix}









\subsection{Data Descriptions}

This study used seven datasets, all of which are intended exclusively for research purposes. None of the datasets are intended for real-world clinical deployment, and we strictly follow their stated usage restrictions.

\begin{itemize}
    \item \textbf{MedQA}: MedQA \cite{jin2021disease} is a large-scale medical question answering dataset constructed from professional medical licensing examinations, including the United States Medical Licensing Examination (USMLE). The dataset contains multiple-choice questions spanning diverse clinical disciplines such as internal medicine, surgery, pathology, pharmacology, and physiology.

    \item \textbf{PubMedQA}: PubMedQA \cite{jin2019pubmedqa} is a biomedical QA dataset collected from PubMed abstracts. Each sample consists of a research question, a biomedical abstract context, a long-form answer derived from the article conclusion, and a corresponding yes/no/maybe label. The dataset contains expert-annotated, unlabeled, and automatically generated QA pairs centered on biomedical research literature.

    \item \textbf{MedMCQA}: MedMCQA \cite{pal2022medmcqa} is a large-scale multiple-choice medical QA dataset collected from Indian medical entrance and licensing examinations such as AIIMS and NEET. The dataset covers numerous medical specialties and includes metadata such as subject categories, explanations, and difficulty-related information for each question.

    \item \textbf{BioASQ}: BioASQ \cite{krithara2023bioasq} is a biomedical benchmark developed as part of the BioASQ challenge, containing factoid, list, yes/no, and summary-style questions grounded in PubMed literature. The dataset includes expert-curated answers together with supporting biomedical articles and is designed to reflect realistic biomedical information-seeking scenarios.

    \item \textbf{MedXpertQA}: MedXpertQA \cite{zuo2025medxpertqa} is a comprehensive expert-level medical benchmark spanning 17 medical specialties and 11 body systems. The study uses the text-only subset of MedXpertQA. Questions are sourced from specialty board examinations and expert-reviewed for accuracy and difficulty.

    \item \textbf{MMLU-Med}: MMLU-Med \cite{hendryckstest2021} is a medical-domain subset of the Massive Multitask Language Understanding (MMLU) benchmark composed of multiple-choice questions from subjects such as anatomy, clinical knowledge, genetics, nutrition, professional medicine, and college-level biology. The dataset covers both foundational biomedical sciences and clinical medicine topics.

    \item \textbf{MedQuAD}: MedQuAD \cite{BenAbacha-BMC-2019} is a medical QA dataset collected from question-answer pairs available on websites of the U.S. National Institutes of Health (NIH). The dataset contains consumer health questions and authoritative answers related to diseases, symptoms, medications, treatments, and healthcare procedures across multiple medical domains.
\end{itemize}

The detailed dataset split and sample counts are shown in Table \ref{tab:data_splits}. Training, validation and testing all consist of a blend of MCQ and open-ended questions.

\begin{table}[h]
\centering
\small
\setlength{\tabcolsep}{4pt}
\renewcommand{\arraystretch}{1.2} 
\begin{tabularx}{\linewidth}{lllc}
\toprule
Data Split & Dataset & Type & \# QA Pairs \\
\midrule

\multirow{4}{*}{Train}
& MedQA     & MCQ        & 10178  \\
& PubMedQA  & MCQ        & 800    \\
& MedMCQA   & MCQ        & 91411 \\
& BioASQ    & Open-ended & 167   \\
\midrule

\multirow{4}{*}{Validation}
& MedQA     & MCQ        & 1272 \\
& PubMedQA  & MCQ        & 100  \\
& MedMCQA   & MCQ        & 1046 \\
& BioASQ    & Open-ended & 42  \\
\midrule

\multirow{4}{*}{\shortstack[l]{Testing\\(Internal)}}
& MedQA     & MCQ        & 1273 \\
& PubMedQA  & MCQ        & 100  \\
& MedMCQA   & MCQ        & 2091 \\
& BioASQ    & Open-ended & 325 \\
\midrule

\multirow{3}{*}{\shortstack[l]{Testing\\(External)}}
& MedXpertQA & MCQ        & 2450 \\
& MMLU-Med   & MCQ        & 1111 \\
& MedQuAD    & Open-ended & 1641 \\
\bottomrule
\end{tabularx}
\caption{Dataset splits used in the study.}
\label{tab:data_splits}
\end{table}

\begin{figure}[h]
    \centering
    \includegraphics[width=\linewidth]{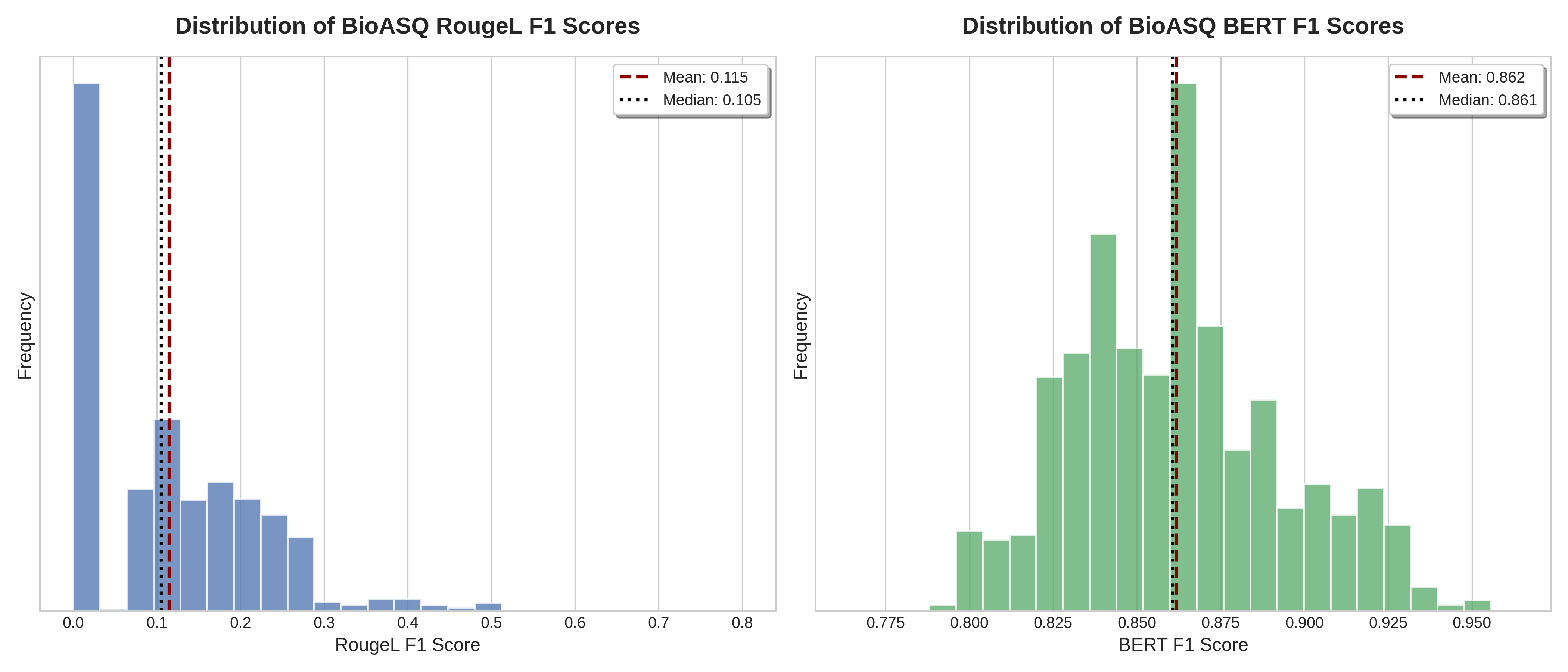}
    \caption{Distribution of ROUGE-L scores on BioASQ predictions. We set the binary correctness threshold at 0.11 (median), which also coincides with a clear separation between two modes.}
    \label{fig:bioasq_hist}
\end{figure}

We use BioASQ as the open-ended QA dataset for PRM training and validation. To derive a binary correctness label, we threshold the ROUGE-L score at 0.11. This value is both the empirical median and aligns with a visible separation between two clusters in the score distribution (Figure~\ref{fig:bioasq_hist}).

\section{Detailed Performance}

Tables~\ref{tab:detail_res_id}, \ref{tab:detail_res_ood}, and \ref{tab:cluster_compare_full} report the complete per-dataset and per-backbone results for the in-distribution benchmarks, the out-of-distribution benchmarks, and the comparison between reasoning trees constructed with and without semantic abstraction, respectively. All experiments are conducted over five random seeds, and the mean and standard deviation are reported. The PRM configurations used in these tables, as well as in the aggregated results presented in the main text, use $\delta_{\mathrm{sim}}=0.80$ and $\alpha=0$, such that the supervision signal is defined solely by proximity to the nearest correct solution.

\newcommand{\stdpm}[1]{\scalebox{0.72}{\ensuremath{\pm#1}}}
\begin{table*}[t]
\centering
\scriptsize
\setlength{\tabcolsep}{2.5pt}
\renewcommand{\arraystretch}{0.95}
\resizebox{\textwidth}{!}{%
\begin{tabular}{llccccccccc}
\toprule
\multirow{3}{*}{Backbone LLMs} & \multirow{3}{*}{Inf. Method}
& \multicolumn{5}{c}{In-distribution} & \multicolumn{4}{c}{Out-of-distribution} \\
\cmidrule(lr){3-7}\cmidrule(lr){8-11}
& & MedQA & MedMCQA & PubMedQA & \multicolumn{2}{c}{BioASQ}
& MedXpertQA & MMLU-Med & \multicolumn{2}{c}{MedQuAD} \\
\cmidrule(lr){6-7}\cmidrule(lr){10-11}
& & Acc $\uparrow$ & Acc $\uparrow$ & Acc $\uparrow$ & BERTScore $\uparrow$ & RougeL $\uparrow$
& Acc $\uparrow$ & Acc $\uparrow$ & BERTScore $\uparrow$ & RougeL $\uparrow$ \\
\midrule
\multirow{4}{*}{Qwen3-8B}
& Best-of-N              & 58.5\stdpm{1.3} & 58.3\stdpm{6.1} & 35.3\stdpm{4.9} & 85.5\stdpm{2.3} & 16.2\stdpm{0.3} & 12.4\stdpm{2.5} & 73.0\stdpm{2.0} & 82.7\stdpm{0.6} & 12.3\stdpm{0.3} \\
& Best-of-N (uncluster)  & 58.0\stdpm{2.1} & 55.0\stdpm{1.3} & 30.5\stdpm{3.5} & 85.5\stdpm{0.2} & 15.8\stdpm{1.1} & 12.7\stdpm{0.0} & 74.1\stdpm{3.7} & 79.6\stdpm{4.2} & 12.5\stdpm{0.5} \\
& MCTS                   & 56.2\stdpm{1.0} & 58.6\stdpm{3.2} & 52.7\stdpm{1.5} & 88.9\stdpm{0.2} & 26.7\stdpm{1.5} & 11.1\stdpm{0.6} & 74.7\stdpm{1.7} & 84.2\stdpm{0.8} & 7.4\stdpm{0.1} \\
& MCTS (uncluster)       & 54.5\stdpm{0.9} & 58.0\stdpm{1.0} & 54.0\stdpm{1.0} & 88.0\stdpm{0.9} & 26.6\stdpm{0.4} & 13.1\stdpm{2.3} & 73.5\stdpm{1.1} & 84.5\stdpm{0.3} & 6.8\stdpm{0.9} \\
\cmidrule(lr){1-11}
\multirow{4}{*}{Llama-3.1-8B-Instruct}
& Best-of-N              & 55.2\stdpm{1.0} & 51.4\stdpm{3.9} & 58.3\stdpm{4.7} & 86.9\stdpm{0.4} & 11.6\stdpm{1.2} & 11.4\stdpm{2.0} & 76.5\stdpm{1.1} & 86.0\stdpm{1.0} & 15.7\stdpm{0.2} \\
& Best-of-N (uncluster)  & 54.2\stdpm{2.1} & 53.8\stdpm{1.4} & 55.3\stdpm{2.5} & 85.6\stdpm{1.5} & 10.9\stdpm{0.4} & 12.7\stdpm{4.5} & 75.8\stdpm{4.5} & 85.2\stdpm{1.7} & 16.0\stdpm{0.5} \\
& MCTS                   & 50.8\stdpm{2.8} & 46.4\stdpm{4.6} & 55.5\stdpm{3.1} & 87.2\stdpm{0.5} & 18.8\stdpm{3.0} & 10.8\stdpm{3.1} & 74.1\stdpm{2.2} & 82.2\stdpm{0.2} & 4.5\stdpm{0.6} \\
& MCTS (uncluster)       & 48.5\stdpm{1.5} & 42.9\stdpm{1.0} & 58.0\stdpm{3.6} & 87.2\stdpm{0.4} & 18.8\stdpm{1.8} & 11.1\stdpm{4.4} & 68.4\stdpm{0.4} & 82.2\stdpm{0.4} & 4.7\stdpm{0.1} \\
\cmidrule(lr){1-11}
\multirow{4}{*}{gemma-2-9b-it}
& Best-of-N              & 52.8\stdpm{2.5} & 40.5\stdpm{2.7} & 23.0\stdpm{5.3} & 74.4\stdpm{0.8} & 16.9\stdpm{0.7} & 8.5\stdpm{3.0} & 67.0\stdpm{2.7} & 73.9\stdpm{4.7} & 9.3\stdpm{0.4} \\
& Best-of-N (uncluster)  & 50.8\stdpm{2.6} & 42.3\stdpm{2.7} & 26.3\stdpm{4.7} & 77.6\stdpm{3.8} & 15.2\stdpm{1.2} & 6.5\stdpm{2.3} & 72.3\stdpm{3.3} & 71.4\stdpm{3.1} & 9.4\stdpm{0.8} \\
& MCTS                   & 54.3\stdpm{0.9} & 45.3\stdpm{1.5} & 33.5\stdpm{2.6} & 87.3\stdpm{1.2} & 24.8\stdpm{1.9} & 9.6\stdpm{2.2} & 69.3\stdpm{2.5} & 80.8\stdpm{3.1} & 3.5\stdpm{0.5} \\
& MCTS (uncluster)       & 55.8\stdpm{0.8} & 48.3\stdpm{2.9} & 37.0\stdpm{4.0} & 87.5\stdpm{1.0} & 27.3\stdpm{2.0} & 11.1\stdpm{0.6} & 68.0\stdpm{3.0} & 81.4\stdpm{0.4} & 3.8\stdpm{0.3} \\
\cmidrule(lr){1-11}
\multirow{4}{*}{Olmo-3-7B-Instruct}
& Best-of-N              & 30.3\stdpm{3.6} & 38.1\stdpm{6.0} & 27.3\stdpm{0.6} & 80.8\stdpm{1.9} & 11.9\stdpm{1.7} & 8.5\stdpm{2.8} & 45.4\stdpm{4.2} & 73.2\stdpm{5.4} & 11.6\stdpm{0.8} \\
& Best-of-N (uncluster)  & 31.2\stdpm{2.4} & 37.5\stdpm{1.9} & 27.3\stdpm{3.1} & 74.6\stdpm{4.0} & 12.0\stdpm{0.7} & 10.8\stdpm{1.7} & 51.6\stdpm{3.7} & 56.9\stdpm{6.7} & 8.8\stdpm{0.8} \\
& MCTS                   & 40.0\stdpm{2.0} & 40.8\stdpm{3.2} & 39.5\stdpm{3.9} & 85.3\stdpm{1.7} & 17.1\stdpm{0.8} & 12.7\stdpm{4.1} & 59.8\stdpm{2.2} & 81.8\stdpm{0.7} & 4.8\stdpm{0.4} \\
& MCTS (uncluster)       & 40.8\stdpm{0.3} & 37.2\stdpm{1.4} & 40.7\stdpm{2.1} & 83.3\stdpm{0.7} & 16.0\stdpm{2.0} & 8.5\stdpm{0.6} & 58.5\stdpm{3.2} & 81.1\stdpm{1.7} & 4.2\stdpm{0.3} \\
\cmidrule(lr){1-11}
\multirow{4}{*}{Phi-3.5-mini-instruct}
& Best-of-N              & 36.2\stdpm{1.5} & 46.5\stdpm{2.1} & 11.7\stdpm{1.5} & 76.6\stdpm{2.2} & 5.3\stdpm{0.5} & 10.1\stdpm{2.3} & 57.0\stdpm{2.2} & 73.1\stdpm{3.8} & 6.9\stdpm{1.0} \\
& Best-of-N (uncluster)  & 38.8\stdpm{14.7} & 45.0\stdpm{8.6} & 11.3\stdpm{1.5} & 79.0\stdpm{4.7} & 6.2\stdpm{1.3} & 8.2\stdpm{4.5} & 55.6\stdpm{23.3} & 69.1\stdpm{19.5} & 6.9\stdpm{1.9} \\
& MCTS                   & 47.3\stdpm{1.2} & 47.4\stdpm{2.8} & 44.7\stdpm{3.2} & 87.9\stdpm{0.3} & 19.3\stdpm{1.5} & 6.5\stdpm{2.0} & 65.2\stdpm{1.3} & 78.9\stdpm{1.8} & 4.0\stdpm{0.4} \\
& MCTS (uncluster)       & 49.0\stdpm{1.5} & 45.0\stdpm{3.1} & 48.7\stdpm{2.3} & 87.1\stdpm{0.4} & 22.6\stdpm{3.2} & 9.5\stdpm{1.5} & 64.9\stdpm{3.7} & 80.6\stdpm{1.8} & 4.0\stdpm{0.3} \\
\bottomrule
\end{tabular}%
}

\caption{Ablation on semantic clustering for Best-of-N and MCTS.}
\label{tab:cluster_compare_full}
\end{table*}

\section{Implementation Details}

We collect offline reasoning trajectories from three open-weight LLMs: Qwen3-8B, Gemma-2-9B-IT, and Llama-3.1-8B-Instruct. For each question, we sample $N=200$ solutions per model using fixed decoding parameters: temperature $0.7$, top-$p=0.9$, repetition penalty $1.1$, and maximum generation length of $8192$ tokens. Each output is formatted as a sequence of atomic reasoning steps marked by \texttt{<|step|>}, followed by a final answer marked by \texttt{<|answer|>}.

Each reasoning step is embedded with the sentence transformer \texttt{all-MiniLM-L6-v2}. We then cluster step embeddings by cosine similarity to construct candidate nodes in the abstract reasoning tree.

The clustering abstraction process takes equivalent of around 300 GPU hours on an NVIDIA H100 GPU, and the inference takes additional 100 GPU hours. The BERTScore reported is based on \texttt{distilroberta-base}. All performance numbers reported in Table \ref{tab:main_results} are the averages across five individual runs with different seeds. Performances in Figure \ref{fig:grid}, Figure \ref{fig:gap_loss} and Figure \ref{fig:scale_mcts} are from singular runs.

More implementation details can be found at \href{https://github.com/zhongys97/MedCoast}{https://github.com/zhongys97/MedCoast}. All nine trained PRMs are released on Huggingface \href{https://huggingface.co/zhongys97/MedCoast}{https://huggingface.co/zhongys97/MedCoast}.

\subsection{Shared Prefix Pairs}

Using the averaged node score in Equation~\ref{eq:score_weight}, we qualitatively examine whether the score captures meaningful differences between alternative continuations from the same reasoning prefix. Table~\ref{tab:gap_siblings_example} shows examples consisting of the original question, the shared prefix, and the diverging next-step candidates from sibling child nodes. The pairs are sampled from the evaluation split using trees constructed with clustering threshold $\delta_{\mathrm{sim}}=0.8$.

For interpretability, we report the correctness and proximity components for each child node. During training, these components are combined into a single node score according to the weighting parameter $\alpha$, and the child with the higher score is assigned the positive label while the lower-scoring child is assigned the negative label. Overall, larger score gaps tend to correspond to more substantial semantic differences between sibling continuations.

\subsection{Hard Negative Pairs Building}
\label{sec:hard_neg_sampling}

We define a gap-dependent acceptance function

\begin{equation}
p(x) \;=\; \lambda \;+\; (1-\lambda)\,\exp\!\left(-\frac{x}{\tau}\right),
\label{eq:gap_sampling}
\end{equation}

where $x \ge 0$ denotes the score gap between two candidates. The parameter $\lambda \in (0,1)$ specifies a minimum acceptance probability, ensuring that samples with large gaps are still retained with non-zero probability. The temperature parameter $\tau > 0$ controls the rate of exponential decay: smaller values of $\tau$ bias the sampling toward smaller gaps, while larger values yield a smoother, more uniform selection. In our implementation, we picked $\tau=0.6$ and $\lambda=0.3$.

\subsection{Critic Training and Inference}

The step-level PRM is initialized from ModernBERT~\cite{warner2025smarter}. We freeze the first 18 layers during training, resulting in 20M trainable parameters out of 149M total parameters. The PRM is optimized with AdamW using a learning rate of $2 \times 10^{-5}$ and batch size $128$. We save checkpoints every 2000 steps and apply early stopping based on validation performance; training is terminated if no improvement is observed over the most recent 50 checkpoints. The scale of data required for PRM training is listed in Table \ref{tab:data_scale}.

\begin{table}[t]
\centering
\small
\begin{tabularx}{\columnwidth}{@{}Xr@{}}
\toprule
\textbf{Quantity} & \textbf{Count} \\
\midrule
Training and validation questions                              & $\sim$105K \\
Generated trajectories                          & $\sim$63M  \\
Retained trajectories (all retained)            & $\sim$63M  \\
Clustered reasoning trees (one per question)    & $\sim$105K \\
Available PRM training pairs                    & $\sim$68M  \\
PRM training pairs used until convergence       & $\sim$15M  \\
\bottomrule
\end{tabularx}
\caption{Scale of offline construction and PRM training. PRM training pairs are positive--negative sibling pairs.}
\label{tab:data_scale}
\end{table}

At inference time, all methods are evaluated under a fixed generation budget. For greedy decoding, beam search, and MCTS, we control the number of candidate expansions $K$, beam size $C$, maximum search depth $S$, and number of simulations $B$ to ensure comparable total model calls across methods. Results in Table \ref{tab:main_results} and Figure \ref{fig:scale_mcts} use the PRM with subtree correctness ration $\alpha=0$ and clustering threshold $\delta_{\mathrm{sim}}=0.80$.

\subsection{Direct Evaluation MCTS}
\label{appendix:mcts}

\paragraph{State and action representation.}
A state $s = (x_1, \dots, x_t)$ is an ordered tuple of committed reasoning step strings.
Each step is delimited by a special \texttt{<|step|>} token prepended by the backbone.
An action $a \in \mathcal{A}(s)$ is a candidate next step string.
A state is terminal if its last step contains \texttt{<|answer|>} or if $|s| \geq D_{\text{search}}$,
where $D_{\text{search}}$ is the maximum committed-step depth. In our implementation, $D_{\text{search}}=20$.

\paragraph{Expansion.}
When a node $s$ is visited for the first time, we query the backbone $\pi_\theta$ with $K$
parallel requests, each conditioned on the concatenation of all steps in $s$ and a
continuation prompt that does not force a final answer.
The $K$ responses are deduplicated via exact-string matching and cached as
$\mathcal{A}(s)$; the cached set is reused across all subsequent simulations and
across committed steps.
MCTS statistics ($N_s$, $N_{sa}$, $W_{sa}$) are reset at the start of each
committed-step search, but the expansion cache is preserved.

\paragraph{Selection.}
At each interior node we select the action that maximises the PUCT score:
\begin{equation}
\mathrm{PUCT}(s, a)
  = Q(s,a)
  + c_{\mathrm{puct}}\,\frac{1}{|\mathcal{A}(s)|}\,
    \frac{\sqrt{N_s + \epsilon}}{1 + N_{sa}},
\label{eq:puct}
\end{equation}
where $Q(s,a) = W_{sa}/N_{sa}$ (0 if $N_{sa}=0$), $N_s$ is the visit count of node $s$,
$N_{sa}$ is the visit count of edge $(s,a)$, $\epsilon = 10^{-8}$, and
$1/|\mathcal{A}(s)|$ is a uniform prior over actions at node $s$.
Node visit counts $N_s$ are incremented during the forward (selection) pass;
edge statistics $N_{sa}$ and $W_{sa}$ are updated during backup. Across all MCTS runs, we fix $c_{\mathrm{puct}}=1.2$ to balance exploration and exploitation.

\paragraph{Evaluation and backup.}
When the simulation reaches a node $s'$ that is either terminal, at the depth limit, or newly expanded leaf, the critic $V_\phi$ is queried once to obtain a scalar value $v = V_\phi(Q, s')$. The value is backed up along the selected path: for each $(s, a)$ in the path, $N_{sa} \leftarrow N_{sa}+1$ and $W_{sa} \leftarrow W_{sa}+v$.

\paragraph{Action commitment.}
After $B$ simulations, the root action is chosen by
$a^* = \arg\max_{a \in \mathcal{A}(s_0)} N_{s_0 a}$,
i.e.\ the most-visited child.
The search then re-roots at $s_0 + (a^*)$ and runs $B$ fresh simulations for
the next committed step.
This continues for at most $D_{\text{tree}}$ committed steps.

\paragraph{Answer forcing.}   
If no committed step contains \texttt{<|answer|>} after the search terminates,
the full reasoning prefix is passed to the backbone with an answer-forcing prompt
that requires the model to emit a final answer.
If the answer token is still absent, the prompt is applied a second time.

\subsection{Token Count}

The inference token count for tree-search methods refers to the total number of generated tokens across all search expansions, including intermediate generations that are later discarded but cached during search. Consequently, the total inference token count is typically much larger than the number of tokens appearing in the final visible response. In Table~\ref{tab:main_results}, the 4096-token budget is applied to this total inference token count rather than the surface response length. For closed-source models, token counts are estimated using the tokenizer of Llama-3.1-8B-Instruct for consistency across APIs.

\subsection{Prompts}

We use a minimal set of prompts for both trajectory collection and inference. Tables~\ref{tab:fr_collect_prompt} and~\ref{tab:mcq_collect_prompt} show the prompts used to collect trajectories for open-ended and multiple-choice questions, respectively. During inference, the prompt in Table~\ref{tab:inf_start_prompt} elicits the first reasoning step, while the prompt in Table~\ref{tab:inf_step_prompt} generates subsequent steps conditioned on the current reasoning prefix. If the maximum step budget of $20$ is reached before an explicit answer is produced, we use the answer-forcing prompt in Table~\ref{tab:inf_ans_prompt} to obtain a final decision.

\section{Semantic Abstraction of the Offline Tree}

Table~\ref{tab:tree_stats} shows that the overall tree structure is not highly sensitive to the inter-cluster similarity threshold $\delta_{\mathrm{sim}}$. As expected, the total number of nodes increases with larger thresholds, since stricter similarity requirements reduce merging and produce more clusters from the same set of reasoning steps. Consistent with this finer-grained partitioning, the average branching factor decreases as the threshold increases, indicating that the tree becomes deeper or more fragmented rather than substantially broader.

\begin{table}[t]
\centering
\small
\setlength{\tabcolsep}{3pt}
\renewcommand{\arraystretch}{1.2} 
\begin{tabular}{lccc}

\hline
\textbf{Metric} & $\delta_{\mathrm{sim}}=0.75$ & \textbf{$\delta_{\mathrm{sim}}=0.80$} & \textbf{$\delta_{\mathrm{sim}}=0.85$} \\
\hline
Tree size & $1532 \pm 769$ & $1924 \pm 898$ & $2402 \pm 1021$ \\
Branch factor & $6.93 \pm 1.86$ & $6.85 \pm 1.59$ & $6.57 \pm 1.29$ \\
Num. leaves & $210 \pm 71$ & $273 \pm 96$ & $357 \pm 112$ \\
Avg. leaf depth & $10.55 \pm 1.65$ & $9.75 \pm 1.47$ & $8.94 \pm 1.26$ \\
Depth & $19.91 \pm 2.03$ & $19.91 \pm 2.03$ & $19.91 \pm 2.03$ \\
\hline
\end{tabular}
\caption{Tree statistics under different clustering threshold similarities.}
\label{tab:tree_stats}
\end{table}

Beyond these trends, the tree statistics reveal substantial compression of the offline reasoning space. The average branching factor remains between approximately 6.5 and 7 across thresholds. Starting from $600$ trajectories sampled from three backbone models, with each trajectory containing roughly $10$ reasoning steps on average, the construction compresses thousands of granular step instances into fewer than $2400$ nodes in a single reasoning tree. This suggests that many superficially different reasoning steps collapse into a smaller set of recurring semantic states.

Tables~\ref{tab:cluster_ex1} and~\ref{tab:cluster_ex2} provide qualitative examples of clusters produced with $\delta_{\mathrm{sim}}=0.75$. The examples show clear intra-cluster semantic coherence as well as meaningful inter-cluster distinctions. For example, in Table~\ref{tab:cluster_ex1}, steps in Cluster 1 primarily discuss the connection between endometrial cancer and Lynch syndrome, whereas Cluster 2 focuses on genetic predisposition. Cluster 3 also concerns genetic factors, but frames them as support for genetic testing rather than as background scientific knowledge. Cluster 4 emphasizes the role of family history of Lynch syndrome, with less focus on mutation mechanisms or prevalence. Table~\ref{tab:cluster_ex2} shows child clusters from the root node, illustrating that different clusters often approach the same question from distinct reasoning perspectives.

\section{Hyperparameter Sensitivity}

\paragraph{Critic ranking performance.}
Table~\ref{tab:val_acc_cutoff} reports the pairwise classification accuracy of the PRM on the validation set, using the best checkpoint before early stopping. Across the nine PRM configurations, two trends emerge. First, the node-scoring weight $\alpha$ has a clear effect on ranking accuracy: subtree-correctness-only supervision ($\alpha=1$) yields the lowest accuracies, whereas proximity-based supervision ($\alpha=0$) yields the highest. This suggests that progress toward a correct answer provides a more learnable signal for distinguishing sibling continuations than empirical subtree correctness alone. Second, the semantic clustering threshold $\delta_{\mathrm{sim}}$ has a comparatively smaller effect, indicating that the learned critic is relatively robust to moderate changes in tree granularity.

\begin{table*}[t]
\centering
\small
\setlength{\tabcolsep}{4pt}
\renewcommand{\arraystretch}{1.2} 
\begin{tabular}{ccccccccc}
\hline
\textbf{$\delta_{\mathrm{sim}}$} & \textbf{$\alpha$} & \textbf{Avg\_acc} & \textbf{Acc ($>$0.1)} & \textbf{Acc ($>$0.2)} & \textbf{Acc ($>$0.3)} & \textbf{Acc ($>$0.4)} & \textbf{Acc ($>$0.5)} & \textbf{Acc ($>$0.6)} \\
\hline
\multirow{3}{*}{0.75}
& 0.0 & 59.95 & 60.48 & 62.08 & 63.64 & 64.31 & 64.37 & 64.47 \\
& 0.5 & 57.57 & 57.79 & 57.65 & 58.29 & 60.57 & 61.14 & 64.57 \\
& 1.0 & 52.25 & 52.42 & 52.47 & 52.30 & 53.27 & 54.93 & 58.50 \\
\hline
\multirow{3}{*}{0.80}
& 0.0 & 60.35 & 61.20 & 63.30 & 65.15 & 65.22 & 64.68 & 64.39 \\
& 0.5 & 55.90 & 55.81 & 55.63 & 58.83 & 60.82 & 60.59 & 64.67 \\
& 1.0 & 51.43 & 51.36 & 51.49 & 51.14 & 52.53 & 54.67 & 59.15 \\
\hline
\multirow{3}{*}{0.85}
& 0.0 & 60.93 & 61.84 & 64.36 & 66.30 & 66.21 & 65.25 & 63.78 \\
& 0.5 & 54.11 & 53.64 & 53.14 & 57.98 & 58.73 & 59.67 & 62.43 \\
& 1.0 & 51.01 & 50.98 & 50.92 & 50.86 & 51.44 & 54.08 & 55.68 \\
\hline
\end{tabular}
\caption{Validation accuracy (\%) under different node-clustering thresholds $\delta_{\mathrm{sim}}$ and subtree-correctness weights $\alpha$. Higher $\alpha$ places greater emphasis on empirical subtree correctness, while lower $\alpha$ emphasizes progress toward the nearest correct solution.}
\label{tab:val_acc_cutoff}
\end{table*}

\paragraph{Downstream inference performance.}
\label{appendix:inf_sensitivity}

\begin{figure*}
    \centering
    \includegraphics[width=\linewidth]{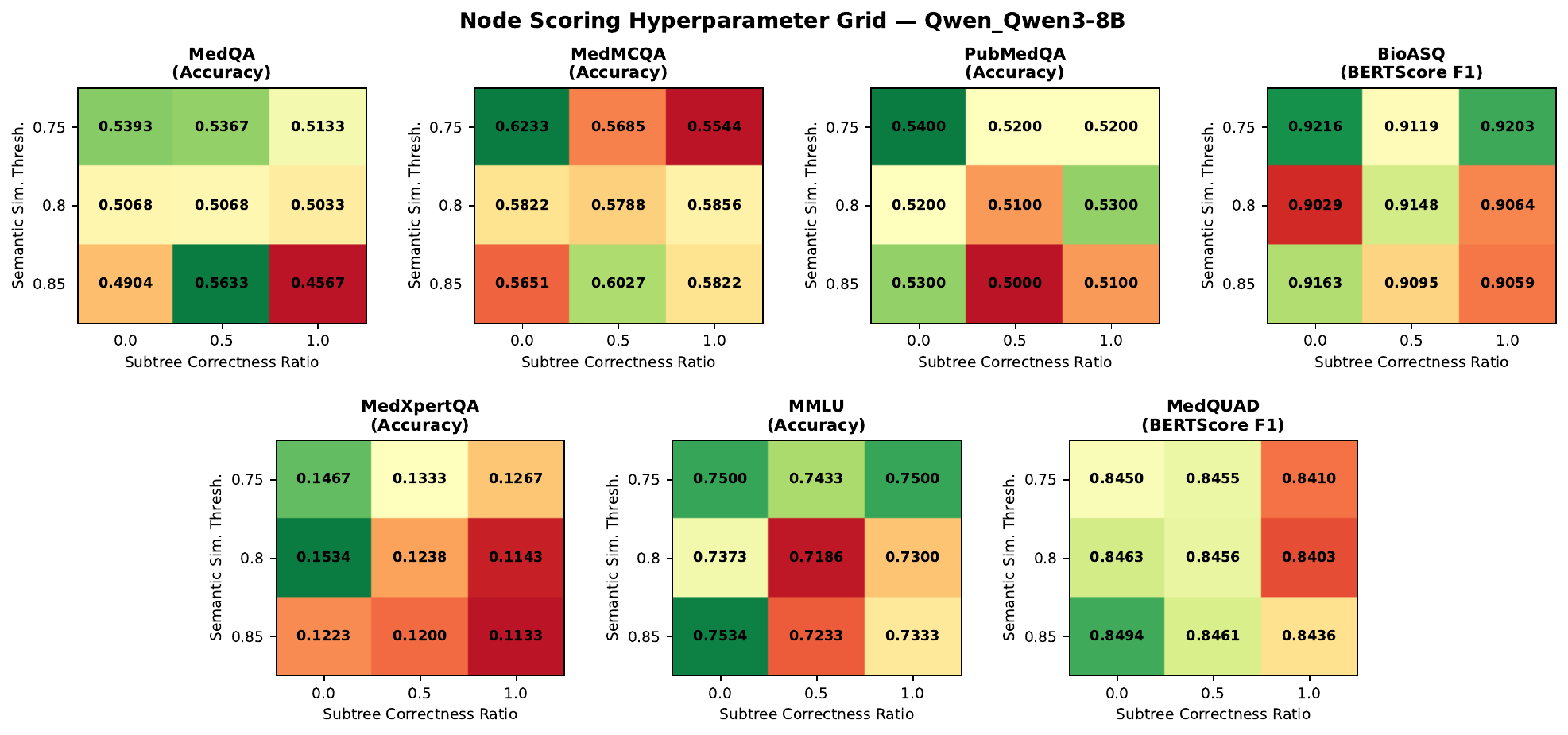}
    \caption{Grid-view ablation of PRM node-scoring design across all seven datasets using Qwen3-8B with direct-evaluation MCTS ($B=16$, $K=16$). Rows vary the semantic clustering threshold $\delta_{\mathrm{sim}}$, and columns vary the subtree-correctness weight $\alpha$, where lower $\alpha$ emphasizes progress toward the nearest correct solution.}
    \label{fig:grid_qwen}
\end{figure*}

\begin{figure*}
    \centering
    \includegraphics[width=\linewidth]{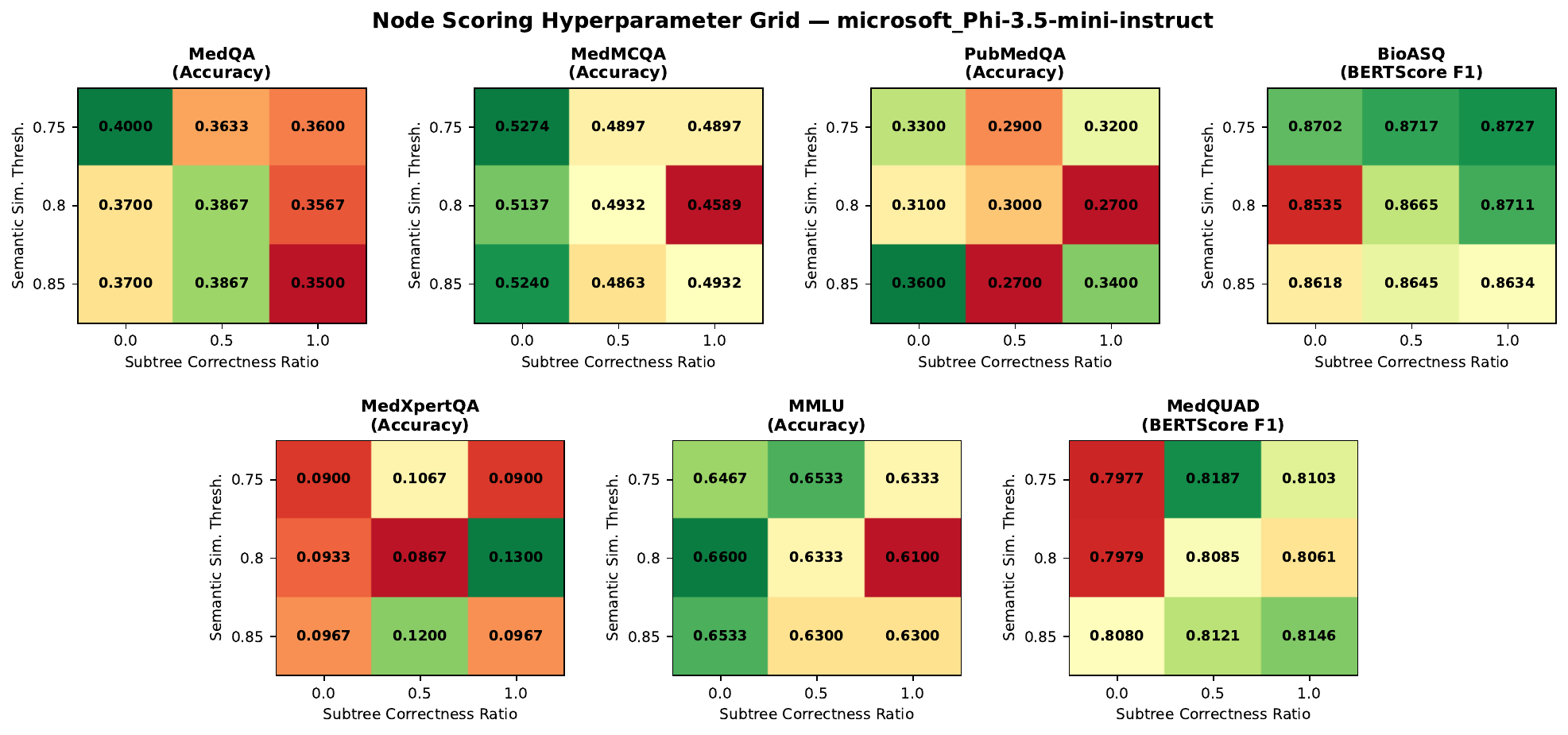}
    \caption{Grid-view ablation of PRM node-scoring design across all seven datasets using Phi-3.5-mini-Instruct with direct-evaluation MCTS ($B=8$, $K=4$).}
    \label{fig:grid_phi}
\end{figure*}

Figures~\ref{fig:grid_qwen} and~\ref{fig:grid_phi} report downstream ablations of the node-scoring design for Qwen3-8B and Phi-3.5-mini, respectively. These results complement the OLMo-3-7B grid-view ablation in Figure~\ref{fig:grid}.

From Table~\ref{tab:prm-sensitivity}, neither PRM training hyperparameter has a statistically significant effect on downstream QA performance for either model (all $p > 0.19$). However, the subtree-correctness weight $\alpha$ appears to be the more important design parameter. Across both backbones, $\alpha$ produces a larger marginal effect ($\sigma = 0.58$ for Qwen3-8B and $0.68$ for Phi-3.5-mini) than the clustering cutoff ($\sigma = 0.46$ and $0.37$, respectively), as well as consistently higher $F$-statistics across experimental settings. This suggests that downstream performance is more sensitive to how node scores balance subtree correctness and progress than to moderate changes in the semantic clustering threshold used during tree construction. Although neither factor reaches statistical significance, the combined analysis indicates that $\alpha$ is the relatively more influential hyperparameter.

\begin{table}[h]
\centering                                                                                                                        \setlength{\tabcolsep}{4pt}
\renewcommand{\arraystretch}{1.2} 
\small
\begin{tabular}{lcccccc}
\toprule
\multirow{2}{*}{Model}
& \multicolumn{2}{c}{Effect Size ($\sigma$)}
& \multicolumn{2}{c}{$\delta_{\mathrm{sim}}$}
& \multicolumn{2}{c}{$\alpha$} \\
\cmidrule(lr){2-3}\cmidrule(lr){4-5}\cmidrule(lr){6-7}
& $F$ & $p$ & $F$ & $p$ & $F$ & $p$ \\
\midrule
Qwen3-8B        & 0.464 & 0.582 & 1.21 & .463 & 2.72 & .325 \\
Phi-3.5-mini    & 0.367 & 0.679 & 0.67 & .636 & 3.85 & .198 \\
\midrule
Combined & 0.415 & 0.631 & 0.94 & .549 & 3.28 & .262 \\
\bottomrule
\end{tabular}
\caption{PRM hyperparameter sensitivity analysis aggregated across all seven datasets. Effect size ($\sigma$) is the standard deviation of marginal means after z-score normalisation per dataset. $F$-statistics are one-way tests treating the other factor as replicates, averaged over datasets. No factor reaches significance at conventional threshold.}
\label{tab:prm-sensitivity}
\end{table}

\section{Time Efficiency}

Table~\ref{tab:compute_resources} reports the wall-clock time and hardware required for trajectory collection, semantic clustering, PRM training, and inference. We acknowledge that constructing the PRM requires a substantial one-time offline investment, corresponding to approximately 300 GPU-hours. Nevertheless, PRM training converged after processing less than 5\% of the available training pairs in the first epoch, equivalent to approximately 15 million pairs.

For context, the process- and reward-model training costs of MedS3~\cite{jiang2025meds} and Med-PRM~\cite{yun2025med} can be estimated to be on the order of 200 GPU-hours based on their reported training configurations. 
\begin{table}[t]
\centering
\small
\setlength{\tabcolsep}{8pt}
\renewcommand{\arraystretch}{1.15}
\begin{tabular}{p{0.58\linewidth}cc}
\toprule
\textbf{Tasks (all models combined)}
& \textbf{Hours}
& \textbf{\# H100} \\
\midrule
Trajectory sampling
& 8
& 20 \\

Step clustering
& 12
& 10 \\

Reward model training
& 15
& 1 \\

Inference (MCTS: 8 simulations, 8 branches, search depth 6)
& 15
& 10 \\
\bottomrule
\end{tabular}
\caption{Computational resources required for each stage of \our.}
\label{tab:compute_resources}
\end{table}

\section{Qualitative Analyses}

\subsection{Human Evaluation Guideline}
\label{appendix:human_eval_guide}

The human evaluation guideline in Table~\ref{tab:human_guide} defines four possible error types: repetition, knowledge gap, reasoning error, and failure to provide a final answer.

\begin{table}[t]
\centering
\small
\begin{tcolorbox}[
  enhanced,
  breakable,
  title={Human Evaluation Guideline},
  colback=white,
  colframe=black,
  boxrule=0.6pt,
  arc=2pt,
  left=6pt,right=6pt,top=6pt,bottom=6pt,
  before skip=6pt, after skip=6pt
]

Goal: Identify common problems in model-generated medical reasoning. For each sample, read the question, reasoning trace, and final answer, then select all applicable issue types. Multiple labels may be selected. Choose ``No Issue'' only if none of the other issues apply. Aim for 3--4 minutes per sample.\newline

\textbf{Repetition}: The model repeats the same idea across steps without meaningful progress, even if phrased differently.

\textbf{Knowledge Gap}: The model makes a medically or factually incorrect claim, such as an incorrect drug, anatomy, diagnosis, mechanism, or treatment. Select only when confident.

\textbf{Reasoning Error}: The facts appear plausible, but the model draws an unsupported or illogical conclusion.

\textbf{No Answer}: The response lacks a clear final answer, or the answer is incomplete or cut off.

\textbf{No Issue}: The reasoning is coherent, medically plausible, non-repetitive, and ends with a clear answer.

\end{tcolorbox}
\caption{Human evaluation guideline.}
\label{tab:human_guide}
\end{table}

Two authors participated in the human evaluation. Each author independently annotated the sampled responses using the guideline above. We report the union of their annotations: a response is considered to exhibit an issue if either annotator flagged that issue. The counts reported in Table~\ref{tab:human_analysis} are computed after this union operation.

\subsection{Examples of Correct Answers}

The correctly answered examples in Table \ref{tab:qa_examples_correct} show that the model can often produce coherent and task-aligned reasoning traces when the required medical knowledge is available. Across these cases, the reasoning typically identifies the key concept in the question, narrows the relevant clinical or biomedical mechanism, and then maps that evidence to the final answer. For instance, the model correctly connects tailored interventions with improved mammography uptake, recognizes Q fever as a non–lice-transmitted infection, identifies essential fatty acids, and associates coarse facial features with Hurler syndrome.

\subsection{Examples of Problematic Answers}
\label{appendix:wrong}

Tables~\ref{tab:fail_rep},~\ref{tab:fail_miss_ans},~\ref{tab:fail_knowledge_gap}, and~\ref{tab:fail_trace} present representative examples of the four abnormal behaviors observed among incorrect predictions: repetitive reasoning, missing final answers, medical knowledge gaps, and failure to select the best answer despite partially relevant reasoning. In each example, the problematic spans are highlighted in red or brown to indicate where the failure occurs within the generated trace.

Table~\ref{tab:fail_rep} provides a particularly direct illustration of how repetitive reasoning can inflate the token budget without adding new evidence or moving the model closer to a final answer. The model repeatedly revisits the same clinical point across multiple steps, which creates the appearance of extended reasoning but contributes little additional decision-making value. This behavior is important for our setting because \our is evaluated under fixed inference budgets: tokens spent on redundant reasoning reduce the amount of computation available for exploring genuinely useful continuations.

\subsection{CoT vs MedCoast}

To understand how CoT and \our would progress from the same prefix, we present a few example cases in Table \ref{tab:cot_mcts_qualitative}. Across the three examples, \our produces more cautious and clinically grounded reasoning than standard CoT. In Example 1, MCTS avoids prematurely ruling out relevant information and instead considers several management strategies, including bowel rest, fluid resuscitation, and monitoring, which helps guide subsequent reasoning steps. In Example 2, the CoT trajectory contains evidence that could support both options A and C but ultimately selects A without sufficiently considering the role of the submucosa. MCTS, in contrast, explicitly reasons about how the options differ and arrives at a better-supported conclusion. A similar pattern appears in Example 3: CoT gives a broad explanation that the anatomical structure is complex and therefore prone to abnormalities, whereas MCTS supplements this observation with more specific disease possibilities.

\subsection{Other Examples}

Table \ref{tab:rep_q_cot} shows the exact backbone (Llama-3.1-8B-Instruct) answering the same two questions in Table \ref{tab:fail_rep} under single pass Chain-of-Thought setup. Notably, the circular reasoning pattern is also observed in Chain-of-Thought in Example 2. Example 1 shows borderline duplicated reasoning as well, though on the surface the model is ruling out options.

\section{Acknowledgments}

The authors acknowledge the AI Makerspace of the College
of Engineering (RRID:SCR 028058), provided by the Partnership for an Advanced Computing Environment (PACE)
at the Georgia Institute of Technology, Atlanta, Georgia,
USA (RRID:SCR 027619), for computational resources used
in this work.
We also gratefully acknowledge funding and
fellowships that contributed to this work, including a Wallace H. Coulter
Distinguished Faculty Fellowship, a Petit Institute Faculty
Fellowship, and research funding from Amazon and Microsoft
Research awarded to Professor May D. Wang.

AI assistants supported language polishing and debugging; authors designed methods, ran experiments, verified results, and reviewed all final claims. In compliance with the institutional rule, the Qwen model used in the study was run outside of PACE.

\begin{table*}[t]
\centering
\small
\providecommand{\stdpm}[1]{\,{\tiny$\pm$#1}}
\renewcommand{\arraystretch}{1.05}
\providecommand{\ours}{\cellcolor{gray!12}}
\begin{tabular*}{\textwidth}{@{\extracolsep{\fill}}lllccccc@{}}
\toprule
\multirow{2}{*}{Backbone LLMs} & \multirow{2}{*}{Reward Model} & \multirow{2}{*}{Search}
& MedQA & MedMCQA & PubMedQA & \multicolumn{2}{c}{BioASQ} \\
\cmidrule(lr){7-8}
& & & Acc $\uparrow$ & Acc $\uparrow$ & Acc $\uparrow$ & BERTScore $\uparrow$ & RougeL $\uparrow$ \\
\midrule
\multirow{7}{*}{Qwen3-8B}
& --                                       & Self-Cons.       & 53.4\stdpm{0.3} & 56.3\stdpm{1.0} & 46.0\stdpm{3.1} & 86.5\stdpm{1.3} & 12.8\stdpm{0.5} \\
\addlinespace[3pt]
&                                          & Best-of-N        & \textbf{61.2}\stdpm{0.6} & \textbf{60.1}\stdpm{2.6} & 36.0\stdpm{3.0} & 83.6\stdpm{1.2} & 17.2\stdpm{0.6} \\
& \multirow{-2}{*}{MedS3}                  & MCTS             & 55.3\stdpm{2.8} & 57.7\stdpm{0.9} & \underline{52.3}\stdpm{3.5} & 88.2\stdpm{1.0} & \textbf{27.8}\stdpm{2.3} \\
\addlinespace[3pt]
&                                          & Best-of-N        & 57.7\stdpm{2.4} & 57.4\stdpm{1.4} & 31.3\stdpm{1.5} & 83.7\stdpm{1.5} & 16.4\stdpm{1.0} \\
& \multirow{-2}{*}{Med-PRM}                & MCTS             & 54.8\stdpm{1.1} & 55.9\stdpm{2.5} & 51.0\stdpm{1.4} & \textbf{89.1}\stdpm{0.5} & \underline{27.7}\stdpm{3.1} \\
\addlinespace[3pt]
& \ours                                    & \ours Best-of-N  & \ours \underline{58.5}\stdpm{1.3} & \ours 58.3\stdpm{6.1} & \ours 35.3\stdpm{4.9} & \ours 85.5\stdpm{2.3} & \ours 16.2\stdpm{0.3} \\
& \ours \multirow{-2}{*}{\textbf{MedCoast}} & \ours MCTS       & \ours 56.2\stdpm{1.0} & \ours \underline{58.6}\stdpm{3.2} & \ours \textbf{52.7}\stdpm{1.5} & \ours \underline{88.9}\stdpm{0.2} & \ours 26.7\stdpm{1.5} \\
\midrule
\multirow{7}{*}{Llama-3.1-8B-Instruct}
& --                                       & Self-Cons.       & 55.1\stdpm{1.9} & \underline{58.7}\stdpm{0.9} & 57.0\stdpm{1.2} & 75.0\stdpm{0.8} & 13.7\stdpm{0.2} \\
\addlinespace[3pt]
&                                          & Best-of-N        & \textbf{64.2}\stdpm{0.6} & \textbf{59.2}\stdpm{1.4} & 60.0\stdpm{3.0} & 84.0\stdpm{1.8} & 11.3\stdpm{0.8} \\
& \multirow{-2}{*}{MedS3}                  & MCTS             & 49.2\stdpm{1.8} & 45.3\stdpm{4.1} & \textbf{61.0}\stdpm{3.5} & 86.9\stdpm{0.5} & \underline{18.0}\stdpm{2.9} \\
\addlinespace[3pt]
&                                          & Best-of-N        & \underline{55.7}\stdpm{2.6} & 58.6\stdpm{2.4} & 51.7\stdpm{5.1} & 84.4\stdpm{0.9} & 10.9\stdpm{0.8} \\
& \multirow{-2}{*}{Med-PRM}                & MCTS             & 50.5\stdpm{4.2} & 43.2\stdpm{2.5} & \underline{60.5}\stdpm{0.7} & \textbf{87.3}\stdpm{0.1} & 17.9\stdpm{0.0} \\
\addlinespace[3pt]
& \ours                                    & \ours Best-of-N  & \ours 55.2\stdpm{1.0} & \ours 51.4\stdpm{3.9} & \ours 58.3\stdpm{4.7} & \ours 86.9\stdpm{0.4} & \ours 11.6\stdpm{1.2} \\
& \ours \multirow{-2}{*}{\textbf{MedCoast}} & \ours MCTS       & \ours 50.8\stdpm{2.8} & \ours 46.4\stdpm{4.6} & \ours 55.5\stdpm{3.1} & \ours \underline{87.2}\stdpm{0.5} & \ours \textbf{18.8}\stdpm{3.0} \\
\midrule
\multirow{7}{*}{gemma-2-9b-it}
& --                                       & Self-Cons.       & 36.0\stdpm{7.0} & 36.6\stdpm{0.5} & 21.0\stdpm{1.0} & 75.9\stdpm{4.3} & 18.4\stdpm{0.7} \\
\addlinespace[3pt]
&                                          & Best-of-N        & \textbf{56.7}\stdpm{1.5} & 42.0\stdpm{3.4} & 21.0\stdpm{0.0} & 62.8\stdpm{3.0} & 14.8\stdpm{0.5} \\
& \multirow{-2}{*}{MedS3}                  & MCTS             & 53.5\stdpm{0.5} & 44.4\stdpm{2.3} & \textbf{36.7}\stdpm{1.2} & 85.4\stdpm{0.6} & 23.5\stdpm{0.9} \\
\addlinespace[3pt]
&                                          & Best-of-N        & 54.2\stdpm{1.8} & 40.5\stdpm{3.8} & 23.5\stdpm{2.1} & 73.2\stdpm{2.5} & 13.7\stdpm{0.3} \\
& \multirow{-2}{*}{Med-PRM}                & MCTS             & 54.0\stdpm{0.9} & \textbf{45.5}\stdpm{1.9} & \underline{35.0}\stdpm{1.7} & \underline{86.8}\stdpm{0.4} & \textbf{25.8}\stdpm{1.4} \\
\addlinespace[3pt]
& \ours                                    & \ours Best-of-N  & \ours 52.8\stdpm{2.5} & \ours 40.5\stdpm{2.7} & \ours 23.0\stdpm{5.3} & \ours 74.4\stdpm{0.8} & \ours 16.9\stdpm{0.7} \\
& \ours \multirow{-2}{*}{\textbf{MedCoast}} & \ours MCTS       & \ours \underline{54.3}\stdpm{0.9} & \ours \underline{45.3}\stdpm{1.5} & \ours 33.5\stdpm{2.6} & \ours \textbf{87.3}\stdpm{1.2} & \ours \underline{24.8}\stdpm{1.9} \\
\midrule
\multirow{7}{*}{Olmo-3-7B-Instruct}
& --                                       & Self-Cons.       & 33.3\stdpm{1.3} & \textbf{47.5}\stdpm{2.6} & 24.2\stdpm{1.5} & 22.1\stdpm{3.1} & 6.2\stdpm{1.0} \\
\addlinespace[3pt]
&                                          & Best-of-N        & 33.5\stdpm{1.3} & 39.9\stdpm{0.5} & 25.3\stdpm{2.1} & 43.0\stdpm{3.5} & 8.2\stdpm{0.5} \\
& \multirow{-2}{*}{MedS3}                  & MCTS             & \textbf{41.3}\stdpm{0.8} & \underline{40.8}\stdpm{2.3} & \underline{38.3}\stdpm{0.6} & \underline{84.8}\stdpm{1.7} & 15.7\stdpm{1.2} \\
\addlinespace[3pt]
&                                          & Best-of-N        & 30.5\stdpm{3.0} & 32.4\stdpm{4.5} & 21.0\stdpm{5.3} & 42.7\stdpm{3.5} & 7.0\stdpm{0.4} \\
& \multirow{-2}{*}{Med-PRM}                & MCTS             & 39.8\stdpm{3.5} & 39.3\stdpm{0.5} & 35.7\stdpm{1.2} & 84.4\stdpm{1.5} & \textbf{20.3}\stdpm{1.2} \\
\addlinespace[3pt]
& \ours                                    & \ours Best-of-N  & \ours 30.3\stdpm{3.6} & \ours 38.1\stdpm{6.0} & \ours 27.3\stdpm{0.6} & \ours 80.8\stdpm{1.9} & \ours 11.9\stdpm{1.7} \\
& \ours \multirow{-2}{*}{\textbf{MedCoast}} & \ours MCTS       & \ours \underline{40.0}\stdpm{2.0} & \ours \underline{40.8}\stdpm{3.2} & \ours \textbf{39.5}\stdpm{3.9} & \ours \textbf{85.3}\stdpm{1.7} & \ours \underline{17.1}\stdpm{0.8} \\
\midrule
\multirow{7}{*}{Phi-3.5-mini-instruct}
& --                                       & Self-Cons.       & 35.3\stdpm{2.4} & 44.9\stdpm{4.5} & 11.0\stdpm{1.5} & 27.7\stdpm{1.9} & 1.0\stdpm{0.2} \\
\addlinespace[3pt]
&                                          & Best-of-N        & 35.7\stdpm{1.5} & \underline{48.0}\stdpm{0.5} & 12.7\stdpm{2.1} & 42.7\stdpm{0.7} & 4.3\stdpm{0.1} \\
& \multirow{-2}{*}{MedS3}                  & MCTS             & \textbf{49.8}\stdpm{1.9} & 45.3\stdpm{2.8} & \textbf{48.7}\stdpm{1.5} & 87.1\stdpm{1.1} & \textbf{22.2}\stdpm{1.7} \\
\addlinespace[3pt]
&                                          & Best-of-N        & 46.2\stdpm{2.6} & \textbf{48.9}\stdpm{2.9} & 9.0\stdpm{3.0} & 81.6\stdpm{2.2} & 5.5\stdpm{0.8} \\
& \multirow{-2}{*}{Med-PRM}                & MCTS             & \underline{48.3}\stdpm{0.6} & 44.4\stdpm{1.4} & \underline{47.7}\stdpm{1.5} & \underline{87.5}\stdpm{0.8} & \underline{20.6}\stdpm{3.0} \\
\addlinespace[3pt]
& \ours                                    & \ours Best-of-N  & \ours 36.2\stdpm{1.5} & \ours 46.5\stdpm{2.1} & \ours 11.7\stdpm{1.5} & \ours 76.6\stdpm{2.2} & \ours 5.3\stdpm{0.5} \\
& \ours \multirow{-2}{*}{\textbf{MedCoast}} & \ours MCTS       & \ours 47.3\stdpm{1.2} & \ours 47.4\stdpm{2.8} & \ours 44.7\stdpm{3.2} & \ours \textbf{87.9}\stdpm{0.3} & \ours 19.3\stdpm{1.5} \\
\midrule
MedS3 Policy   & MedS3   & Best-of-N & 61.8\stdpm{0.6} & 59.5\stdpm{4.7} & 59.7\stdpm{2.9} & 76.2\stdpm{0.2} & 16.2\stdpm{0.6} \\
Med-PRM Policy & Med-PRM & Best-of-N & 62.2\stdpm{0.3} & 58.3\stdpm{0.7} & 59.0\stdpm{1.7} & 85.1\stdpm{0.9} & 10.5\stdpm{0.3} \\
\bottomrule
\end{tabular*}
\caption{In-distribution results: self-consistency vs.\ Best-of-N and MCTS guided by MedS3, Med-PRM, and MedCoast reward models. MedCoast rows are shaded. Best per backbone in \textbf{bold}, second best \underline{underlined}.}
\label{tab:detail_res_id}
\end{table*}

\begin{table*}[t]
\centering
\small
\providecommand{\stdpm}[1]{\,{\tiny$\pm$#1}}
\renewcommand{\arraystretch}{1.05}
\providecommand{\ours}{\cellcolor{gray!12}}
\begin{tabular*}{\textwidth}{@{\extracolsep{\fill}}lllcccc@{}}
\toprule
\multirow{2}{*}{Backbone LLMs} & \multirow{2}{*}{Reward Model} & \multirow{2}{*}{Search}
& MedXpertQA & MMLU-Med & \multicolumn{2}{c}{MedQuAD} \\
\cmidrule(lr){6-7}
& & & Acc $\uparrow$ & Acc $\uparrow$ & BERTScore $\uparrow$ & RougeL $\uparrow$ \\
\midrule
\multirow{7}{*}{Qwen3-8B}
& --                                       & Self-Cons.       & \textbf{13.3}\stdpm{1.0} & 73.1\stdpm{0.9} & \textbf{87.0}\stdpm{1.4} & \underline{12.5}\stdpm{0.2} \\
\addlinespace[3pt]
&                                          & Best-of-N        & 12.1\stdpm{1.5} & \textbf{86.4}\stdpm{0.9} & 70.2\stdpm{4.4} & 11.1\stdpm{0.8} \\
& \multirow{-2}{*}{MedS3}                  & MCTS             & 11.1\stdpm{1.1} & 74.9\stdpm{1.1} & 84.3\stdpm{0.1} & 6.6\stdpm{0.2} \\
\addlinespace[3pt]
&                                          & Best-of-N        & 10.1\stdpm{2.3} & 83.7\stdpm{1.5} & 80.6\stdpm{2.4} & \textbf{12.7}\stdpm{0.9} \\
& \multirow{-2}{*}{Med-PRM}                & MCTS             & 11.3\stdpm{2.1} & 73.7\stdpm{0.0} & \underline{84.4}\stdpm{0.1} & 6.3\stdpm{0.2} \\
\addlinespace[3pt]
& \ours                                    & \ours Best-of-N  & \ours \underline{12.4}\stdpm{2.5} & \ours 83.0\stdpm{2.0} & \ours 82.7\stdpm{0.6} & \ours 12.3\stdpm{0.3} \\
& \ours \multirow{-2}{*}{\textbf{MedCoast}} & \ours MCTS       & \ours 11.1\stdpm{0.6} & \ours \underline{84.7}\stdpm{1.7} & \ours 84.2\stdpm{0.8} & \ours 7.4\stdpm{0.1} \\
\midrule
\multirow{7}{*}{Llama-3.1-8B-Instruct}
& --                                       & Self-Cons.       & \underline{11.1}\stdpm{0.6} & 68.9\stdpm{1.9} & 83.0\stdpm{0.1} & 14.6\stdpm{0.2} \\
\addlinespace[3pt]
&                                          & Best-of-N        & 8.2\stdpm{1.5} & \textbf{85.4}\stdpm{0.4} & \underline{86.0}\stdpm{1.9} & \underline{15.9}\stdpm{0.2} \\
& \multirow{-2}{*}{MedS3}                  & MCTS             & 10.4\stdpm{3.7} & 70.6\stdpm{0.9} & 82.3\stdpm{0.2} & 4.8\stdpm{0.1} \\
\addlinespace[3pt]
&                                          & Best-of-N        & 10.5\stdpm{2.5} & \underline{77.5}\stdpm{2.4} & \textbf{87.2}\stdpm{0.1} & \textbf{16.4}\stdpm{0.5} \\
& \multirow{-2}{*}{Med-PRM}                & MCTS             & 10.8\stdpm{2.8} & 71.9\stdpm{0.0} & 82.4\stdpm{0.5} & 4.4\stdpm{0.1} \\
\addlinespace[3pt]
& \ours                                    & \ours Best-of-N  & \ours \textbf{11.4}\stdpm{2.0} & \ours 76.5\stdpm{1.1} & \ours \underline{86.0}\stdpm{1.0} & \ours 15.7\stdpm{0.2} \\
& \ours \multirow{-2}{*}{\textbf{MedCoast}} & \ours MCTS       & \ours 10.8\stdpm{3.1} & \ours 74.1\stdpm{2.2} & \ours 82.2\stdpm{0.2} & \ours 4.5\stdpm{0.6} \\
\midrule
\multirow{7}{*}{gemma-2-9b-it}
& --                                       & Self-Cons.       & 3.8\stdpm{1.0} & 53.5\stdpm{0.4} & 46.7\stdpm{1.3} & 6.4\stdpm{2.6} \\
\addlinespace[3pt]
&                                          & Best-of-N        & 7.2\stdpm{0.6} & \underline{79.0}\stdpm{1.1} & 59.5\stdpm{5.3} & 7.9\stdpm{0.7} \\
& \multirow{-2}{*}{MedS3}                  & MCTS             & \textbf{11.1}\stdpm{2.0} & 67.5\stdpm{1.9} & \underline{81.4}\stdpm{0.1} & 3.6\stdpm{0.4} \\
\addlinespace[3pt]
&                                          & Best-of-N        & 6.9\stdpm{2.8} & 75.6\stdpm{0.0} & 70.4\stdpm{3.3} & \textbf{9.7}\stdpm{0.3} \\
& \multirow{-2}{*}{Med-PRM}                & MCTS             & \underline{10.5}\stdpm{1.5} & 68.3\stdpm{1.7} & \textbf{82.3}\stdpm{0.6} & 3.7\stdpm{0.2} \\
\addlinespace[3pt]
& \ours                                    & \ours Best-of-N  & \ours 8.5\stdpm{3.0} & \ours 77.0\stdpm{2.7} & \ours 73.9\stdpm{4.7} & \ours \underline{9.3}\stdpm{0.4} \\
& \ours \multirow{-2}{*}{\textbf{MedCoast}} & \ours MCTS       & \ours 9.6\stdpm{2.2} & \ours \textbf{79.3}\stdpm{2.5} & \ours 80.8\stdpm{3.1} & \ours 3.5\stdpm{0.5} \\
\midrule
\multirow{7}{*}{Olmo-3-7B-Instruct}
& --                                       & Self-Cons.       & 7.5\stdpm{2.5} & \textbf{63.8}\stdpm{1.7} & 33.0\stdpm{2.9} & \underline{5.0}\stdpm{0.8} \\
\addlinespace[3pt]
&                                          & Best-of-N        & 8.2\stdpm{1.1} & 49.1\stdpm{1.5} & 15.4\stdpm{2.5} & 2.3\stdpm{0.6} \\
& \multirow{-2}{*}{MedS3}                  & MCTS             & \underline{12.1}\stdpm{1.5} & 59.0\stdpm{1.7} & 79.3\stdpm{2.4} & 4.0\stdpm{0.4} \\
\addlinespace[3pt]
&                                          & Best-of-N        & 10.5\stdpm{2.5} & 45.7\stdpm{4.1} & 30.5\stdpm{4.6} & 4.0\stdpm{1.0} \\
& \multirow{-2}{*}{Med-PRM}                & MCTS             & 11.8\stdpm{1.0} & \underline{60.5}\stdpm{0.4} & \underline{80.6}\stdpm{2.0} & 4.5\stdpm{0.6} \\
\addlinespace[3pt]
& \ours                                    & \ours Best-of-N  & \ours 8.5\stdpm{2.8} & \ours 45.4\stdpm{4.2} & \ours 73.2\stdpm{5.4} & \ours \textbf{11.6}\stdpm{0.8} \\
& \ours \multirow{-2}{*}{\textbf{MedCoast}} & \ours MCTS       & \ours \textbf{12.7}\stdpm{4.1} & \ours 59.8\stdpm{2.2} & \ours \textbf{81.8}\stdpm{0.7} & \ours 4.8\stdpm{0.4} \\
\midrule
\multirow{7}{*}{Phi-3.5-mini-instruct}
& --                                       & Self-Cons.       & 0.7\stdpm{0.0} & 59.7\stdpm{2.0} & 51.0\stdpm{2.8} & 5.4\stdpm{0.4} \\
\addlinespace[3pt]
&                                          & Best-of-N        & 8.8\stdpm{4.5} & 50.6\stdpm{2.3} & 37.3\stdpm{2.0} & 3.9\stdpm{0.4} \\
& \multirow{-2}{*}{MedS3}                  & MCTS             & \textbf{10.5}\stdpm{0.6} & \underline{65.7}\stdpm{4.5} & \underline{79.0}\stdpm{2.8} & 4.0\stdpm{0.3} \\
\addlinespace[3pt]
&                                          & Best-of-N        & \underline{10.1}\stdpm{3.0} & \textbf{69.1}\stdpm{1.1} & 76.2\stdpm{2.0} & \textbf{8.2}\stdpm{0.8} \\
& \multirow{-2}{*}{Med-PRM}                & MCTS             & 9.8\stdpm{2.6} & 64.2\stdpm{2.6} & \textbf{80.6}\stdpm{0.5} & 3.7\stdpm{0.5} \\
\addlinespace[3pt]
& \ours                                    & \ours Best-of-N  & \ours \underline{10.1}\stdpm{2.3} & \ours 57.0\stdpm{2.2} & \ours 73.1\stdpm{3.8} & \ours \underline{6.9}\stdpm{1.0} \\
& \ours \multirow{-2}{*}{\textbf{MedCoast}} & \ours MCTS       & \ours 6.5\stdpm{2.0} & \ours 65.2\stdpm{1.3} & \ours 78.9\stdpm{1.8} & \ours 4.0\stdpm{0.4} \\
\midrule
MedS3 Policy   & MedS3   & Best-of-N & 9.2\stdpm{1.1} & 80.5\stdpm{1.5} & 73.9\stdpm{2.0} & 12.8\stdpm{0.8} \\
Med-PRM Policy & Med-PRM & Best-of-N & 10.0\stdpm{1.2} & 79.1\stdpm{2.4} & 73.5\stdpm{1.8} & 9.1\stdpm{0.5} \\
\bottomrule
\end{tabular*}
\caption{Out-of-distribution results: self-consistency vs.\ Best-of-N and MCTS guided by MedS3, Med-PRM, and MedCoast reward models. MedCoast rows are shaded. Best per backbone in \textbf{bold}, second best \underline{underlined}.}
\label{tab:detail_res_ood}
\end{table*}

\newcommand{\note}[1]{%
  \colorbox{blue!12}{\strut\footnotesize #1}%
}

\begin{table*}[t]
\centering
\small
\setlength{\tabcolsep}{4pt}
\renewcommand{\arraystretch}{1.12}
\begin{tabular}{p{0.12\linewidth} p{0.85\linewidth}}
\toprule
\textbf{Field} & \textbf{Content} \\
\midrule

Question &
\detokenize{"Women with synchronous primary cancers of the endometrium and ovary: do they have Lynch syndrome?",
  "options": {
    "A": "yes",
    "B": "no",
    "C": "maybe"
  }} \\
\midrule

Prefix &
The presence of synchronous primary cancers in the endometrium and ovary raises suspicion for a hereditary cancer syndrome due to the shared anatomical origin and potential genetic predisposition. \\
\midrule

\multicolumn{2}{c}{\textbf{Cluster 1}} \\
\midrule
&
Lynch syndrome is strongly linked to endometrial cancer, with up to 30\% of endometrial cancers being attributable to Lynch syndrome. \\
\cmidrule(l){2-2}
&
The revised Bethesda guidelines recommend testing for Lynch syndrome in individuals with certain types of cancer, including colorectal, endometrial, ovarian, and other related cancers. \textbackslash{}n\textbackslash{}n For women with synchronous primary cancers of the endometrium and ovary, testing for Lynch syndrome may involve genetic counseling and testing for mutations in the MLH1, MSH2, MSH6, and PMS2 genes. \\
\cmidrule(l){2-2}
&
The lifetime ovarian cancer (OC) risk in Lynch syndrome carriers ranges from 8 - 40\%, but it remains significantly lower than endometrial carcinoma incidence. \\
\cmidrule(l){2-2}
&
Lynch syndrome is the most common hereditary cancer syndrome associated with endometrial and ovarian cancers, particularly when diagnosed at a younger age. \\
\midrule

\multicolumn{2}{c}{\textbf{Cluster 2}} \\
\midrule
&
Exposure to **estrogen therapy** or obesity raises endometrioid ovarian risk separately, mimicking Lynch-associated patterns. These factors could explain synchronous occurrence in the absence of genetic hypermutation. \\
\cmidrule(l){2-2}
&
Synchronous cancers in women suggest a possible genetic predisposition, as these cancers may arise from the same germline mutation.\\
\cmidrule(l){2-2}
&
It also involves increased risk for other adenocarcinomas such as endometrial and ovarian cancers due to germline mutations in DNA mismatch repair genes. \\
\cmidrule(l){2-2}
&
Sporadic synchronous primaries are well-described in postmenopausal women, accounting for up to **10\% of all uterine/ovarian dual malignancies** without inherited syndromes.\\
\midrule

\multicolumn{2}{c}{\textbf{Cluster 3}} \\
\midrule
&
To confirm Lynch syndrome, genetic testing for mismatch repair gene mutations is necessary. \\
\cmidrule(l){2-2}
&
If the genetic testing is positive for a Lynch syndrome-related gene mutation, it would support the diagnosis of Lynch syndrome.\\
\cmidrule(l){2-2}
&
Absence of known germline pathogenic variants does not exclude Lynch; somatic mosaicism exists but is rare compared to classic germline mechanism. \\
\cmidrule(l){2-2}
&
Therefore, a definitive diagnosis of Lynch syndrome requires a comprehensive evaluation, including family history, tumor characteristics, and genetic testing.\\
\midrule

\multicolumn{2}{c}{\textbf{Cluster 4}} \\
\midrule
&
Synchronous cancers in these sites are more commonly linked to Lynch syndrome than to sporadic mutations. \\
\cmidrule(l){2-2}
&
On the other hand, if the patient's family history is negative for Lynch syndrome-associated cancers, or if the tumors do not show evidence of MSI or mismatch repair deficiency, it would decrease the likelihood of Lynch syndrome. \\
\cmidrule(l){2-2}
&
However, it's essential to note that synchronous primary cancers can also occur in the absence of Lynch syndrome, particularly in women with other risk factors, such as family history or genetic mutations. \\
\cmidrule(l){2-2}
&
If the woman has a strong family history of cancer, particularly in first-degree relatives, this would increase the likelihood of Lynch syndrome. \\
\bottomrule
\end{tabular}
\caption{Example of semantic clusters formed during trajectory abstraction. The clustering similarity threshold is $\delta_{\mathrm{sim}}=0.75$.}
\label{tab:cluster_ex1}
\end{table*}

\begin{table*}[t]
\centering
\small
\setlength{\tabcolsep}{4pt}
\renewcommand{\arraystretch}{1.12}
\begin{tabular}{p{0.12\linewidth} p{0.85\linewidth}}
\toprule
\textbf{Field} & \textbf{Content} \\
\midrule

Question &
``Which is the major symptom of the Doose syndrome?'' \\
\midrule

Prefix &
\note{Empty. Just to begin answer the question.} \\
\midrule

\multicolumn{2}{c}{\textbf{Cluster 1}} \\
\midrule
&
Doose syndrome, also known as myoclonic-astatic epilepsy, is a rare genetic disorder that primarily affects the brain's development and function. \\
\cmidrule(l){2-2}
&
Doose syndrome, also known as myoclonic-astatic epilepsy, is a rare genetic disorder that primarily affects the brain's electrical activity. \\
\cmidrule(l){2-2}
&
To determine the major symptom of Doose syndrome, I first need to recall what Doose syndrome is. Doose syndrome is a rare genetic disorder that affects the nervous system. \\
\midrule

\multicolumn{2}{c}{\textbf{Cluster 2}} \\
\midrule
&
The disorder is characterized by seizures that begin in infancy or early childhood. \\
\cmidrule(l){2-2}
&
The condition is characterized by seizures that begin in infancy or early childhood. \\
\cmidrule(l){2-2}
&
The condition is characterized by seizures that typically begin in infancy or early childhood. \\
\midrule

\multicolumn{2}{c}{\textbf{Cluster 3}} \\
\midrule
&
I am looking for the primary clinical presentations in new borns. \\
\midrule

\multicolumn{2}{c}{\textbf{Cluster 4}} \\
\midrule
&
In addition to seizures, patients may experience developmental delays and intellectual disabilities. \\
\cmidrule(l){2-2}
&
In addition to myoclonic seizures, patients may also experience ataxia, which is a lack of coordination and balance. \\
\cmidrule(l){2-2}
&
In addition to seizures and developmental issues, patients may also experience movement disorders such as ataxia. \\
\cmidrule(l){2-2}
&
In addition to seizures, developmental delays and behavioral issues are commonly observed in affected individuals. \\
\bottomrule
\end{tabular}
\caption{Another example of semantic clusters formed during trajectory abstraction. The clustering similarity threshold is $\delta_{\mathrm{sim}}=0.75$.}
\label{tab:cluster_ex2}
\end{table*}

\newcommand{\cscore}[1]{%
  \colorbox{green!15}{\strut\footnotesize Correctness: #1}%
}

\newcommand{\pscore}[1]{%
  \colorbox{blue!12}{\strut\footnotesize Proximity: #1}%
}

\newcommand{\scores}[2]{%
  \hspace{0.4em}\cscore{#1}\hspace{0.3em}\pscore{#2}%
}

\begin{table*}[t]
\centering
\small
\setlength{\tabcolsep}{4pt}
\renewcommand{\arraystretch}{0.9}
\begin{tabular}{p{0.12\linewidth} p{0.85\linewidth}}
\toprule
\textbf{Field} & \textbf{Content} \\
\midrule

\multicolumn{2}{c}{\textbf{Example 1}} \\
\midrule

Question &
"question": "Which of the following is a primarily RNA based technique?",
        "options": {
            "A": "Next generation sequencing",
            "B": "PCR",
            "C": "Sanger's technique",
            "D": "Western blotting"
}
\\
\midrule

Prefix &
<|step|> To determine which technique is primarily RNA based, we must first understand the nature of each technique. Next generation sequencing (NGS) can be used for both DNA and RNA, so it is not primarily RNA based. <|step|> PCR (Polymerase Chain Reaction) is a technique that amplifies DNA, not RNA, so it is not primarily RNA based. <|step|> Sanger's technique is a method for DNA sequencing and is not primarily RNA based.
\\
\midrule
Positive &
Western blotting is a technique used to detect specific proteins, not RNA, so it is not primarily RNA based. \scores{0.47}{0.60}
\\
\midrule
Negative &
Western Blotting detects proteins by binding antibodies to SDS-PAGE separated samples; while cDNA may sometimes derive from RNA transcripts for gene expression analysis. \scores{0.33}{0.20}
\\

\midrule

\multicolumn{2}{c}{\textbf{Example 2}} \\
\midrule

Question &
"question": "A 60-year-old woman comes to the physician because of palpitations and a 5.4-kg (11.9-lb) weight loss over the past 3 months. She appears anxious. Her pulse is 104/min and blood pressure is 148/101 mm Hg. Physical examination shows warm skin, mild tremor, and an enlarged thyroid gland. A thyroid scintigraphy scan shows several focal areas with increased uptake. A biopsy of one of these areas of affected thyroid tissue is most likely to show which of the following?",
        "options": {
            "A": "Sheets of parafollicular cells surrounded by amyloid stroma",
            "B": "Inflammatory infiltrate with multinucleated giant cells and granuloma formation",
            "C": "Follicle destruction with lymphocytic infiltrate and germinal center formation",
            "D": "Patches of enlarged follicular cells distended with colloid"
        }
\\
\midrule

Prefix &
<|step|> The patient presents with symptoms of palpitations, weight loss, and a goiter, which are classic signs of hyperthyroidism. These symptoms suggest overactivity of the thyroid gland. <|step|> The presence of an enlarged thyroid gland and focal areas of increased uptake on thyroid scintigraphy suggests the presence of autonomously functioning thyroid nodules.
\\
\midrule
Positive &
Among the options provided, the one that aligns with an increase in follicular cells is option D: Patches of enlarged follicular cells distended with colloid. This option suggests an increase in the number of follicular cells, which is consistent with a multinodular goiter. \scores{0.67}{1.0}

\\
\midrule
Negative &
However, option D does not match our findings of hyperthyroidism and the scintigraphy pattern of increased uptake. The correct answer should be something that matches these findings. \scores{0.33}{0.33}
\\

\midrule

\multicolumn{2}{c}{\textbf{Example 3}} \\
\midrule

Question &
"question": "A 33-year-old Asian American man presents to his primary care provider in February with pain in his fingers. His symptoms started 2 years ago and have gradually progressed. He initially attributed his symptoms to stress at work as a social worker. He has tried using ibuprofen, acetaminophen, and naproxen, but these medications did not improve his pain. He also reports that there have been at least 3 instances in the past 2 months in which his fingers became extremely painful and turned white before turning pale blue. These instances occurred when the patient was walking outside. His past medical history is notable for gastroesophageal reflux disease and gout. He takes omeprazole and allopurinol. His temperature is 99 F (37.2 C), blood pressure is 120/78 mmHg, pulse is 90/min, and respirations are 18/min. Physical examination reveals loss of skin fold wrinkles and indurated skin on the palmar side of the digits. Finger range of motion is limited. Which of the following medications is most appropriate to address this patient's concerns?",
        "options": {
            "A": "Ambrisentan",
            "B": "Amlodipine",
            "C": "Cilostazol",
            "D": "Lisinopril"
        }
\\
\midrule

Prefix &
<|step|> The patient presents with progressive finger pain and intermittent episodes of finger discoloration, which suggests a vascular or connective tissue disorder. The symptoms are exacerbated by cold exposure, as the episodes occur when walking outside. This is a key clue pointing toward Raynaud's phenomenon or a related condition.
\\
\midrule
Positive &
The patient's laboratory results are not provided, but a diagnosis of limited systemic scleroderma would be supported by the presence of autoantibodies, such as anti-Scl-70 or anti-centromere antibodies. However, in the absence of laboratory results, we can proceed with the clinical diagnosis of limited systemic scleroderma. \scores{0.67}{0.5}

\\
\midrule
Negative &
The patient's symptoms worsen when walking outside, which may suggest cold-induced vasospasm or a systemic condition affecting small vessels. This pattern is not typical of gout or gastroesophageal reflux disease, which are already accounted for in his medical history. \scores{0.33}{0.50}
\\

\bottomrule
\end{tabular}
\caption{QA with parent prefix and positive/negative step candidates and scores with average of percentage correct and progress ratio.}
\label{tab:gap_siblings_example}
\end{table*}

\newcolumntype{L}[1]{>{\arraybackslash}p{#1}}

\begin{table*}[t]
\centering
\small
\setlength{\tabcolsep}{3pt}
\renewcommand{\arraystretch}{1.35}
\begin{tabular}{l L{0.90\linewidth}}
\toprule
\textbf{Field} & \textbf{Content} \\
\midrule

Question
& "Can tailored interventions increase mammography use among HMO women?", "options": \{"A": "yes", "B": "no", "C": "maybe"\} \\

Pred reasoning
& <|step|> The question asks whether tailored interventions can increase mammography use specifically among women in Health Maintenance Organizations (HMOs). Evidence from health promotion studies generally supports that personalized or targeted interventions improve the uptake of recommended screenings among at-risk or specific populations, including those within managed care settings. HMOs often<|step|> Tailored interventions have been shown in literature to increase screening uptake among specific patient groups, including HMO members, by addressing barriers to care and personalizing communication. This aligns with the question's focus on HMO women.<|step|> The question asks whether tailored interventions have sufficient evidence to increase mammography use specifically among HMO-enrolled women. Major reviews and studies in healthcare delivery settings, including HMOs, demonstrate that personalized interventions improve screening adherence by addressing individual barriers and improving provider-patient communication. This evidence base supports the<|answer|>A\\

\midrule

Question
& "Which of the following is not transmitted by lice",
  "options": {
    "A": "Q fever",
    "B": "Trench fever",
    "C": "Relapsing fever",
    "D": "Epidemic typhus"
  }\} \\

Pred reasoning
& <|step|> To determine which disease is not transmitted by lice, I need to identify the vector for each disease. <|step|> Lice are known to transmit trench fever, epidemic typhus, and relapsing fever. Q fever is transmitted by ticks and contaminated environments, not lice.  \textbackslash{}n<|answer|> A \\

\midrule

Question
& "question": "Which of the following is an essential fatty acid?",
  "options": {
    "A": "Linoleic acid",
    "B": "Alpha linolenic acid",
    "C": "Both of the above",
    "D": "None of the above"
  } \\

Pred reasoning
& <|step|> Essential fatty acids (EFAs) are crucial for various bodily functions, including the formation of cell membranes and the production of hormone-like substances called eicosanoids. Linoleic acid and alpha-linolenic acid are the two main EFAs. Linoleic acid is an omega-<|step|> Linoleic acid is an omega-6 fatty acid, which is essential for the human body as it cannot be produced on its own. It plays a crucial role in various bodily functions, including inflammation regulation, skin health, and brain function. <|step|> Both linoleic acid and alpha-linolenic acid are the two main EFAs that the human body cannot synthesize. <|answer|> C \\

\midrule

Question
& "question": "In DIC following are seen except",
                "options": {
                    "A": "Fibrinogen decreased",
                    "B": "Thrombocytopenia",
                    "C": "Normal APTT",
                    "D": "PT elevation"
                } \\

Pred reasoning
& 
<|step|> Identify typical laboratory findings associated with disseminated intravascular coagulation (DIC), which usually includes thrombocytopenia, fibrinogen depletion, prolonged prothrombin time (PT), and activated partial throm<|step|> In normal circumstances during DIC, there would not be an increase in Prothrombin Time (PT); it typically gets extended due to consumption of clotting factors. Therefore, we look at options where PT does not get affected as abnormalities indicative of DIC.<|step|> Reviewing standard lab results expected in DIC; recognize Fibrinogen decrease, Thrombocytopenia, Activated Partial Thromboplastin Time (APTT) extension from factor usage but normally unchanged Prothrombin Time (PT) until later stages<|answer|> C \\

\midrule

Question
& "question": "A one year old child, who is earlier normal, is suffering from copious nasal discharge, coarse facial features, large tongue, bulging head and flattened nose. He is also having enlarged spleen and liver. What may be the cause?",
                "options": {
                    "A": "Hypothyroidism",
                    "B": "Beckwith widman syndrome",
                    "C": "Proteus syndrome",
                    "D": "Hurler syndrome"
                } \\

Pred reasoning
& 
<|step|> Identify symptoms associated with known genetic conditions affecting children such as growth abnormalities, organomegaly, and distinctive physical traits.<|step|> Match presented symptoms including characteristic craniofacial appearance, organomegaly, and developmental issues against descriptions of specific congenital disorders listed among options. The most fitting condition described by these clinical signs aligns with Hurler syndrome.<|step|> Consider all provided symptoms nasal discharge, coarse facial features, macroglossia, prominent forehead, protruding eyes, hepatosplenomegaly and cross-reference them with recognized characteristics of metabolic storage diseases typically present at<|step|> Compare given symptoms with classic presentation of Mucopolysaccharidoses types I through VI; select type corresponding with detailed manifestations outlined.</|step|><|answer|>D\\

\midrule




Question
& "Which transcription factor is considered as a master regulator of lysosomal genes?" \\

Pred reasoning
& <|step|> The master regulator of lysosomal genes is a key transcription factor involved in the regulation of genes related to lysosomal function and biogenesis.  \textbackslash{}n<|answer|> TFEB \\

Notes
& Ground Truth: Transcription Factor EB (TFEB), RougeL-F1: 0.40, BERTscore-F1: 0.843 \\
\bottomrule
\end{tabular}
\caption{Examples of correct answers for both MCQ and open-ended medical questions. Definition of correct answer for open-ended question is RougeL exceeding 0.11.}
\label{tab:qa_examples_correct}
\end{table*}


\begin{table*}[t]
\centering
\small
\setlength{\tabcolsep}{4pt}
\renewcommand{\arraystretch}{1.2}
\begin{tabular}{p{0.12\linewidth} p{0.85\linewidth}}
\toprule
\textbf{Field} & \textbf{Content} \\
\midrule

\multicolumn{2}{c}{\textbf{Example 1}} \\
\midrule

Question &
\detokenize{"question": "Resistance form of endodontics is:",
  "options": {
    "A": "Resists movement of gutta-percha in apical area",
    "B": "To allow use of spreader in lateral condensation",
    "C": "Fracture of root while vertical condensation",
    "D": "None of the above"
  }"}

\\
\midrule
Pred reasoning &
<|step|> The question asks for the definition of the 'resistance form' in endodontics. Resistance form refers to a characteristic of root canal filling that resists the movement of gutta-percha, particularly in the apical area during procedures like vertical condensation.<|step|> The correct description of resistance form is that it resists the movement of gutta-percha in the apical area. This directly matches option A.<|step|> The question asks which option correctly defines the resistance form in endodontics. Based on established endodontic knowledge, the resistance form is specifically described as resisting movement of gutta-percha in the apical area. This matches option A.<|step|> The correct answer must be the option that specifically describes the resistance form in endodontics. Option A states \"Resists movement of gutta-percha in apical area,\" which is the standard definition of resistance form.\red{<|step|> The resistance form in endodontics is defined by its ability to oppose the movement of filling material, specifically gutta-percha, in the apical region during procedures like vertical condensation. This directly aligns with option A.}<|step|> The question provides four options regarding the resistance form in endodontics. Based on dental endodontics knowledge, the resistance form is primarily characterized by its ability to prevent the movement of gutta-percha in the apical area, particularly during procedures like vertical condensation. \red{<|step|> The resistance form in endodontics is known for its role in preventing gutta-percha movement in the apical area. Among the given options, only option A specifically describes this function.<|step|> The resistance form in endodontics is about the physical property that prevents gutta-percha from moving in the apical area, as stated in option A. However, the question specifically asks for the 'form' itself, which in endodontics refers to the anatomical or mechanical resistance provided. <|step|> The resistance form in endodontics is defined by its role in preventing gutta-percha movement in the apical area during filling procedures. Option A directly states this definition. <|step|> The resistance form in endodontics is defined as the property of the root canal system that resists the extrusion or movement of gutta-percha, especially in the apical area during procedures like vertical condensation. Among the options, only option A directly describes this function. }<|step|> The question asks which option correctly defines the resistance form in endodontics. From previous reasoning, the resistance form is known to resist gutta-percha movement in the apical area during root canal procedures, which matches option A.\red{<|step|> The resistance form in endodontics is primarily associated with preventing gutta-percha movement in the apical area during procedures. Among the provided options, only A directly matches this definition.}<|answer|>A
\\
\midrule
GT answer &
\detokenize{D}
\\

\midrule

\multicolumn{2}{c}{\textbf{Example 2}} \\
\midrule

Question &
\detokenize{"question": "How is oprozomib administered?"}
\\
\midrule
Pred reasoning &
<|step|> Oprozomib is administered intravenously. It is typically given as an infusion over 30 to 120 minutes.<|step|> Oprozomib is administered intravenously as an infusion, usually over 30 to 60 minutes per dose.<|step|> Oprozomib is administered via intravenous infusion in a clinical setting.<|step|> Oprozomib is administered as an intravenous infusion, typically in a hospital or outpatient setting.<|step|> The specific route of administration for oprozomib is intravenous infusion.<|step|> The question asks specifically how oprozomib is administered, which from prior steps is known to be intravenously as an infusion.\red{<|step|> Oprozomib is an oral medication, so it is administered by mouth as a tablet or capsule.<|step|> Oprozomib is an oral medication, so it is administered by mouth, not intravenously.<|step|> Oprozomib is an oral medication, so it is administered by mouth.<|step|> The question asks how oprozomib is administered, and from established pharmacology, it is an oral medication.<|step|> Oprozomib is an oral medication, so it is administered by mouth.}<|step|> The question asks how oprozomib is administered. From pharmacology knowledge, oprozomib is an oral medication administered by mouth.<|step|> The question specifically asks how oprozomib is administered. Based on pharmacological knowledge, oprozomib is an oral medication. It is taken by mouth in the form of a tablet or capsule.\red{<|step|> Oprozomib is an oral medication, so it is administered by mouth. It is typically taken as a tablet or capsule.<|step|> Oprozomib is an oral medication, so it is administered by mouth, typically as a tablet or capsule.<|step|> Oprozomib is an oral medication, so it is administered orally by mouth.<|step|> Oprozomib is an oral medication, so it is administered by mouth. It is taken as a tablet or capsule.}<|step|> The question asks how oprozomib is administered. Based on pharmacological knowledge, oprozomib is an oral medication. It is taken by mouth, usually as a tablet or capsule.<|answer|>Oral
\\
\midrule
GT answer &
\detokenize{Orally}
\\

\bottomrule
\end{tabular}
\caption{Representative failure examples caused by repetitive reasoning steps. \red{Red} fonts signal highly similar sentences.}
\label{tab:fail_rep}
\end{table*}

\begin{table*}[t]
\centering
\small
\setlength{\tabcolsep}{4pt}
\renewcommand{\arraystretch}{1.15}
\begin{tabular}{p{0.12\linewidth} p{0.85\linewidth}}
\toprule
\textbf{Category} & \textbf{Content} \\
\midrule
Question &
\detokenize{"question": "Orteronel was developed for treatment of which cancer?"}
\\
\midrule
Pred reasoning &
<|step|> Identify Orteronel as an investigational drug specifically designed against prostate cancer by examining its clinical trial data and research publications focused on hormone therapy agents used in treating advanced stages of malignancies related to male reproductive organs.\textbackslash{}n \red{<|answer}<|step|> Review scientific literature regarding Orteronel's mechanism of action targeting enzymes involved in steroidogenesis within tumor cells specific to prostate carcinoma pathology. This approach helps confirm its developmental purpose without assuming additional information not provided in your query.\textbackslash{}n\red{<|answer}<|step|> Consult authoritative pharmaceutical databases such as PubMed or DrugBank where I would search using keywords \"Orteronel\" along with terms associated with prostate cancer treatments, ensuring my sources directly relate to therapeutic uses rather than general chemistry<|step|> Search through peer-reviewed studies focusing on drugs under investigation for various types of cancer to find mentions of Orteronel being linked explicitly to prostate cancer treatment protocols based on their published outcomes and stated objectives.\textbackslash{}n\red{<|answer|>}\textbackslash{}n\textbackslash{}n
\\
\midrule
GT answer &
\detokenize{B}
\\
\bottomrule
\end{tabular}
\caption{Representative failure examples unable to output the answer.}
\label{tab:fail_miss_ans}
\end{table*}

\begin{table*}[t]
\centering
\small
\setlength{\tabcolsep}{4pt}
\renewcommand{\arraystretch}{1.2}
\begin{tabular}{p{0.12\linewidth} p{0.85\linewidth}}
\toprule
\textbf{Field} & \textbf{Content} \\
\midrule

\multicolumn{2}{c}{\textbf{Example 1}} \\
\midrule

Question &
"question": "A previously healthy 17-year-old girl comes to the emergency department because of a 5-day history of progressive lower abdominal pain, fever, and malodorous vaginal discharge. Menarche was at the age of 12 years, and her last menstrual period was 2 weeks ago. She is sexually active with one male partner and uses a combination contraceptive patch. Her temperature is 37.8 \textdegree C (100\textdegree F), pulse is 90/min, respirations are 22/min, and blood pressure is 110/70 mm Hg. Abdominal examination shows severe lower quadrant tenderness bilaterally. Pelvic examination shows a purulent cervical discharge, cervical motion tenderness, and bilateral adnexal tenderness. Her hemoglobin concentration is 10.5 g/dL, leukocyte count is 13,000/mm3 , and platelet count is 345,000/mm3 . A urine pregnancy test is negative. Which of the following is the most appropriate pharmacotherapy?",
  "options": {
    "A": "Oral azithromycin",
    "B": "Vaginal clindamycin",
    "C": "Intravenous penicillin and vancomycin",
    "D": "Intramuscular ceftriaxone and oral doxycycline"
  }

\\
\midrule
Pred reasoning &
\detokenize{<|step|> The patient presents with lower abdominal pain, fever, malodorous vaginal discharge, history of sexual activity with contraceptive use, and exam findings of pelvic inflammatory disease (PID) with cervical and adnexal tenderness. Lab results show leukocytosis and normal hemoglobin, no pregnancy. <|step|> The presentation and labs suggest a likely bacterial pelvic inflammatory disease (PID), which requires empirical coverage of both Gram-negative and Gram-positive bacteria. Option C provides both beta-lactam (penicillin) and a macrolide (vancomycin), covering these pathogens. <|answer|>C}
\\
\midrule
GT answer &
\detokenize{D}
\\

\midrule

\multicolumn{2}{c}{\textbf{Example 2}} \\
\midrule

Question &
\detokenize{"question": "How many TAp73 isoforms have been identified in humans?"}
\\
\midrule
Pred reasoning &
\detokenize{<|step|> The TAp73 isoforms are a subset of the larger p73 family of proteins, which are involved in regulating cell growth and survival. To determine the number of TAp73 isoforms identified in humans, it is essential to consider the various isoforms generated through alternative splicing and transcription. <|step|> Research indicates that TAp73 is a transcriptional activator isoform of the tumor suppressor protein p73, and it plays a crucial role in promoting cell cycle arrest and apoptosis in response to DNA damage. Studies have identified multiple isoforms of TAp73, primarily generated through alternative splicing<|step|> TAp73 isoforms are generated through alternative splicing and transcriptional regulation. Research on the p73 family has identified multiple isoforms, including TAp73, Np73, and p73, among others. Studies have shown that these isoforms have distinct functions in regulating cell<|answer|> 3}
\\
\midrule
GT answer &
\detokenize{7}

\\

\bottomrule
\end{tabular}
\caption{Representative failure examples caused by medical knowledge gaps.}
\label{tab:fail_knowledge_gap}
\end{table*}

\begin{table*}[t]
\centering
\small
\setlength{\tabcolsep}{4pt}
\renewcommand{\arraystretch}{1.15}
\begin{tabular}{p{0.12\linewidth} p{0.85\linewidth}}
\toprule
\textbf{Category} & \textbf{Content} \\
\midrule
Question &
\detokenize{"question": "A 29-year-old woman presents to the physician with a blurred vision of her right eye for 2 days. She has pain around her right eye during eye movement. She takes no medications. At the clinic, her blood pressure is 110/70 mm Hg, the pulse is 72/min, respirations are 15/min, and the temperature is 36.5 \textdegree (97.7 \textdegree). On physical examination, illumination of the left eye results in bilateral pupillary constriction while illumination of the right eye results in a mild bilateral pupillary dilation. Fundoscopic examination shows optic disk swelling in the right eye. The color vision test shows decreased perception in the right eye. The remainder of the physical examination shows no abnormalities. Specific additional history should be obtained regarding which of the following?",
  "options": {
    "A": "Dry eyes",
    "B": "High-risk sexual behaviour",
    "C": "Oral ulcers",
    "D": "Sensory loss"
  }}
\\
\midrule
Pred reasoning &
<|step|> The patient's symptoms suggest a neurological condition, specifically involving the optic nerve, given the optic disk swelling and altered pupillary response. \red{This points towards a need to investigate for conditions like optic neuritis}, which is often associated with multiple sclerosis. <|step|> Multiple sclerosis is linked to high-risk sexual behavior <|answer|> B
\\
\midrule
GT answer &
\detokenize{D}
\\

\midrule

\multicolumn{2}{c}{\textbf{Example 2}} \\
\midrule

Question &
"question": "A 65-year-old woman is brought to the emergency department by her husband after he found her with nausea, headache, and agitation 1 hour ago. When he left their lakeside cabin 6 hours before to go ice fishing and get more firewood, she did not have any symptoms. She has chronic migraines, hypertension, and type 2 diabetes mellitus. Her medications include lisinopril and metformin, but she ran out of her anti-hypertensive medications over the weekend. Her temperature is 37.1 \textdegree C (98.8 \textdegree F), pulse is 110/min, respirations are 21/min, and blood pressure is 154/92 mm Hg. Pulse oximetry on room air shows an oxygen saturation of 98\%. She is confused and oriented only to person and place. She recalls only one of three objects after 5 minutes. Her gait is unsteady. Physical examination shows no abnormalities. Which of the following is the most effective intervention for this patient's current condition?",
  "options": {
    "A": "Intravenous nitroprusside",
    "B": "Intranasal sumatriptan",
    "C": "Hyperbaric oxygen therapy",
    "D": "100\% oxygen therapy"
  }
\\
\midrule
Pred reasoning &
<|step|> Assess the severity of the patient's acute presentation including confusion, altered mental status, high heart rate, elevated blood pressure without focal neurological deficits. The primary concern here seems related to cerebrovascular insufficiency. <|step|> Given the clinical scenario indicates potential stroke due to sudden change in consciousness, history of hypertension, and physical findings such as confusion and ataxia; prioritize rapid management targeting cerebral perfusion improvement while considering contraindications from existing comorbidities.<|step|> In patients presenting with signs suggestive of transient ischemic attack or minor stroke such as sudden confusion, difficulty speaking, visual disturbances, dizziness, loss of balance, coordination problems, weakness on one side of the body, slurred speech.<|step|> Considering the urgent need to restore adequate cerebral blood flow and minimize brain injury duration, choose between options aiming directly at improving cerebral perfusion.<|step|> Among the given options, intravenous administration allows immediate effectiveness and precise control necessary during acute events involving possible strokes.<|answer|> A
\\
\midrule
GT answer &
\detokenize{C}
\\

\bottomrule
\end{tabular}
\caption{Representative examples due to medically sound reasoning but nearly missing the best answer.}
\label{tab:fail_trace}
\end{table*}


\newcommand{\steptok}{%
  \texttt{\textless|step|\textgreater}%
}
\newcommand{\answertok}{%
  \texttt{\textless|answer|\textgreater}%
}

\begin{table*}[t]
\centering
\small
\setlength{\tabcolsep}{4pt}
\renewcommand{\arraystretch}{0.9}
\begin{tabular}{p{0.12\linewidth} p{0.85\linewidth}}
\toprule
\textbf{Field} & \textbf{Content} \\
\midrule

\multicolumn{2}{c}{\textbf{Example 1}} \\
\midrule

Question &
``question'': ``A 64-year-old male patient on the surgical floor develops new abdominal discomfort and bloating 48 hours after undergoing an uncomplicated right hemicolectomy. He has not passed flatus or stool since the procedure. His vital signs include a heart rate of 110/min, blood pressure of 100/56 mmHg, respiratory rate of 20/min, and temperature of \(37.2^\circ\mathrm{C}\) (99.0~\(^{\circ}\mathrm{F}\)). On examination, his abdomen is distended, mildly tender, tympanic to percussion, and has diminished bowel sounds. A peripheral IV is placed, and IV fluids are administered. What is the most appropriate next step in managing this patient?'', ``options'': ``A'': ``Initiate total parenteral nutrition'', ``B'': ``Administer opioid analgesics for pain relief'', ``C'': ``Examine the surgical wounds'', ``D'': ``Measure serum lactate levels'', ``E'': ``Obtain an abdominal X-ray'', ``F'': ``Perform an exploratory laparotomy'', ``G'': ``Start broad-spectrum antibiotics'', ``H'': ``Administer prokinetic agents'', ``I'': ``Insert a nasogastric tube'', ``J'': ``Perform a CT scan of the abdomen and pelvis''
\\
\midrule

Prefix &
\steptok\ The patient presents with post-operative ileus characterized by absent bowel sounds, abdominal distention, and inability to pass gas or stool.
\\
\midrule

CoT next step &
\steptok\ The immediate priority is to rule out intestinal obstruction which could necessitate urgent intervention.
\\
\midrule

MCTS next step &
\steptok\ Initial management of postoperative ileus typically involves supportive measures such as bowel rest, fluid resuscitation, and monitoring.
\\
\midrule

\multicolumn{2}{c}{\textbf{Example 2}} \\
\midrule

Question &
``question'': ``The mucosa of the hard palate is:'', ``options'': ``A'': ``Keratinised and has submucosa and minor salivary glands posterolaterally'', ``B'': ``Non-keratinised and has submucosa and minor salivary glands posteromedially'', ``C'': ``Keratinised and lacks submucosa and minor salivary glands'', ``D'': ``Non-keratinised and lacks submucosa and minor salivary glands''
\\
\midrule

Prefix &
\steptok\ Identify the type of epithelium found on the surface of the hard palate based on anatomical knowledge.
\\
\midrule

CoT next step &
\steptok\ The majority of the human mouth's interior surfaces where food contact occurs need protection against mechanical stress; hence they would likely exhibit keratinized tissue characteristics. \answertok\ A
\\
\midrule

MCTS next step &
\steptok\ The majority of the human mouth's surfaces including the hard palate exhibit keratinized stratified squamous epithelium due to its exposure to mechanical stress from food manipulation during chewing. This tissue also contains underlying connective tissues such as submucosa.
\\
\midrule

\multicolumn{2}{c}{\textbf{Example 3}} \\
\midrule

Question &
``question'': ``Highest incidence of atypical root canal or root among all teeth is seen in:'', ``options'': ``A'': ``Mandibular 1st premolar'', ``B'': ``Mandibular canine'', ``C'': ``Maxillary 1st premolar'', ``D'': ``Mandibular 2nd premolar''
\\
\midrule

Prefix &
\steptok\ The question asks about the highest incidence of atypical root canal or root among all teeth. \steptok\ To answer this, we need to consider the anatomical characteristics and common complications associated with each tooth mentioned in the options. \steptok\ Among the options provided, the mandibular first premolar is known to have a prominent groove and a small lingual cusp.
\\
\midrule

CoT next step &
\steptok\ Considering the anatomical characteristics and common complications associated with each tooth, the mandibular first premolar is known to have a complex root canal system with curved and narrow canals, which increases the likelihood of atypical root canal anatomy, such as a calcified canal or an additional canal.
\\
\midrule

MCTS next step &
\steptok\ The mandibular first premolar is known to have a unique root canal anatomy, with a high incidence of bifurcations, accessory canals, and potential for calcification. Additionally, the tooth is commonly associated with complications such as root resorption, periodontal disease, and pulp cavity.
\\

\bottomrule
\end{tabular}
\caption{Qualitative comparison of the next reasoning step produced by chain-of-thought (CoT) decoding and MCTS from the same prefix.}
\label{tab:cot_mcts_qualitative}
\end{table*}


\begin{table*}[t]
\centering
\small
\setlength{\tabcolsep}{4pt}
\renewcommand{\arraystretch}{1.2}
\begin{tabular}{p{0.12\linewidth} p{0.85\linewidth}}
\toprule
\textbf{Field} & \textbf{Content} \\
\midrule

\multicolumn{2}{c}{\textbf{Example 1}} \\
\midrule

Question &
\detokenize{"question": "Resistance form of endodontics is:",
  "options": {
    "A": "Resists movement of gutta-percha in apical area",
    "B": "To allow use of spreader in lateral condensation",
    "C": "Fracture of root while vertical condensation",
    "D": "None of the above"
  }"}

\\
\midrule
Pred reasoning &
<|step|> To determine the correct answer, let's first understand the concept of resistance form in endodontics. In endodontics, resistance form refers to the ability of the root canal filling material to resist the movement or displacement of the gutta-percha in the apical area. This is crucial for maintaining the integrity of the apical seal and preventing the migration of bacteria or other substances into the root canal.\textbackslash{}n\textbackslash{}n<|step|> The options provided can be analyzed as follows: option A suggests that resistance form is related to resisting the movement of gutta-percha in the apical area, which aligns with the definition of resistance form. Option B mentions the use of a spreader in lateral condensation, but this is not directly related to the concept of resistance form. Option C describes a fracture of the root while using vertical condensation, which is not the primary function of resistance form.\textbackslash{}n\textbackslash{}n<|step|> Considering the above analysis, option A appears to be the most relevant to the concept of resistance form in endodontics. However, we should confirm whether it is the correct answer by examining the relationship between resistance form and gutta-percha movement.\textbackslash{}n\textbackslash{}n<|step|> The resistance form is indeed associated with the resistance of the gutta-percha in the apical area, but it is more accurately described as the ability of the root canal filling material to resist the vertical and lateral forces that can displace the gutta-percha. This is a critical aspect of maintaining the apical seal.\textbackslash{}n\textbackslash{}n<|step|> Based on this understanding, option A is a reasonable description of resistance form in endodontics, but it is essential to consider if it is the most comprehensive or accurate option.\textbackslash{}n\textbackslash{}n<|step|> Upon further consideration, it seems that option A provides a concise description of resistance form, but it may not capture the full scope of its implications. However, without additional information, it appears to be the most suitable option.\textbackslash{}n\textbackslash{}n<|answer|> A
\\
\midrule
GT answer &
\detokenize{D}
\\

\midrule

\multicolumn{2}{c}{\textbf{Example 2}} \\
\midrule

Question &
\detokenize{"question": "How is oprozomib administered?"}
\\
\midrule
Pred reasoning &
<|step|> The question is about the administration of oprozomib, a proteasome inhibitor used in the treatment of certain types of cancer. To determine how it is administered, we need to consider the typical methods of administration for proteasome inhibitors.\textbackslash{}n\textbackslash{}n \red{Proteasome inhibitors are usually administered intravenously (IV) or orally. Given that oprozomib is a relatively new medication, it is likely that its administration method would follow this trend.\textbackslash{}n\textbackslash{}n However, to confirm the specific administration method for oprozomib, we need to consult reliable sources such as the medication's labeling, clinical trial data, or reputable medical literature.} \shade{[..SKIP FEW STEPS.]} <|step|> The oprozomib medication label or package insert is not readily available, and clinical trial data or research articles do not provide sufficient information on the administration method. In this case, we can proceed to the next step to explore other sources.\textbackslash{}n\textbackslash{}n \brown{However, based on the general trend of proteasome inhibitors being administered IV or orally, it is reasonable to infer that oprozomib may be administered via one of these routes.}\textbackslash{}n\textbackslash{}n <|step|> The administration method of oprozomib is not explicitly stated in the provided information. However, considering the typical administration methods for proteasome inhibitors, it is likely that oprozomib is administered intravenously (IV) or orally.\textbackslash{}n\textbackslash{}n To confirm this inference, we would need more specific information on the medication's labeling, clinical trial data, or research articles.\textbackslash{}n\textbackslash{}n\red{<|step|> Considering the general trend of proteasome inhibitors being administered IV or orally, and the lack of specific information on oprozomib's administration method, it is reasonable to infer that oprozomib may be administered intravenously (IV) or orally.\textbackslash{}n\textbackslash{}n However, this inference is based on limited information, and a more definitive answer would require access to the medication's labeling, clinical trial data, or research articles.}\textbackslash{}n\textbackslash{}n\brown{<|step|> In the absence of specific information on oprozomib's administration method, it is possible that the medication may be administered via a different route. However, based on the general trend of proteasome inhibitors, it is reasonable to infer that oprozomib may be administered intravenously (IV) or orally.\textbackslash{}n\textbackslash{}n To confirm this inference, we would need more specific information on the medication's labeling, clinical trial data, or research articles.{<|step|> Oprozomib is a proteasome inhibitor used in the treatment of certain types of cancer.} \red{Proteasome inhibitors are typically administered intravenously (IV) or orally. However, the specific administration method for oprozomib is not explicitly stated in the provided information.\textbackslash{}n\textbackslash{}n Considering the typical administration methods for proteasome inhibitors, it is reasonable to infer that oprozomib may be administered intravenously (IV) or orally.}\textbackslash{}n\textbackslash{}n<|answer|> Oprozomib is likely administered intravenously (IV) or orally.}\textbackslash{}n \textbackslash{}n Concise final answer: Oprozomib is likely administered intravenously (IV) or orally.
\\
\midrule
GT answer &
\detokenize{Orally}
\\

\bottomrule
\end{tabular}
\caption{One-pass Chain-of-Thought responses for the same example questions shown in Table~\ref{tab:fail_rep}. Example 1 is verbose but does not exhibit clear looping, whereas Example 2 repeatedly returns to the same concept without reaching a resolution. Colored fonts indicate repeated semantics.}
\label{tab:rep_q_cot}
\end{table*}

\begin{table*}[t]
\centering
\small
\begin{tcolorbox}[
  enhanced,
  breakable,
  title={Multiple-Choice Question Collection System Prompt},
  colback=white,
  colframe=black,
  boxrule=0.6pt,
  arc=2pt,
  left=6pt,right=6pt,top=6pt,bottom=6pt,
  before skip=6pt, after skip=6pt
]
You are a medical expert answering domain specific questions. \newline

    Your goal is to take small, reasoning that move toward the correct answer.\newline

    STRICT RULES\newline
    - Produce a sequence of steps (3 - 20).\newline
    - Each step MUST represent exactly ONE justification.\newline
    - Each step MUST be around (20 - 30 words).\newline
    - Answer with maximum budget of 20 steps.\newline
    - Do NOT repeat earlier steps.\newline
    - Do NOT assume missing patient facts.\newline
    - Use medically correct knowledge only.\newline
    
    PROCESS
    - Prefer discriminative reasoning: use findings to eliminate options, not broad explanations.\newline
    - If the stem lacks key data, explicitly name the missing piece(s) and proceed with the most reasonable inference from what IS given.\newline
    - Commit to an option only after you have ruled out alternatives or identified a defining criterion.\newline

    FORMAT
    - Each step MUST begin with <|step|>.\newline
    - The final answer (if reached) must begin with <|answer|>.\newline
    - If the answer is not yet determined, continue with another <|step|>.\newline

    MULTIPLE CHOICE QUESTIONS
    - Do NOT choose an option until justified by prior steps.\newline
    - When choosing, use <|answer|> followed by ONLY the option letter.\newline

    UNCERTAINTY
    - If information is insufficient, add a step stating what is missing.\newline
    - Still proceed with the most reasonable medical inference.\newline
\end{tcolorbox}
\caption{Multiple-Choice Question Collection System Prompt}
\label{tab:mcq_collect_prompt}
\end{table*}

\begin{table*}[t]
\centering
\small
\begin{tcolorbox}[
  enhanced,
  breakable,
  title={Open-ended Question Collection System Prompt},
  colback=white,
  colframe=black,
  boxrule=0.6pt,
  arc=2pt,
  left=6pt,right=6pt,top=6pt,bottom=6pt,
  before skip=6pt, after skip=6pt
]

    You are a medical expert answering domain specific questions.\newline

    Your goal is to take small, reasoning that move toward the correct answer.\newline

    STRICT RULES\newline
    - Produce a sequence of steps (3 - 20).\newline
    - Each step MUST represent exactly ONE justification.\newline
    - Each step MUST be around (20 - 30 words).\newline
    - Answer with maximum budget of 20 steps.\newline
    - Do NOT repeat earlier steps.\newline
    - Do NOT assume missing patient facts.\newline
    - Use medically correct knowledge only.\newline

    FORMAT\newline
    - Each step MUST begin with <|step|>.\newline
    - The final answer (if reached) must begin with <|answer|>.\newline
    - After <|answer|>, give a concise final answer (less than 40 tokens).\newline
    - If the answer is not yet determined, continue with another <|step|>.\newline
    - Do NOT include <|step|> after <|answer|>.\newline

    UNCERTAINTY\newline
    - If information is insufficient, add a step stating what is missing.\newline
    - Still proceed with the most reasonable medical inference.\newline
\end{tcolorbox}
\caption{Open-ended Question Collection System Prompt}
\label{tab:fr_collect_prompt}
\end{table*}

\begin{table*}[t]
\centering
\small
\begin{tcolorbox}[
  enhanced,
  breakable,
  title={Inference Start Prompt},
  colback=white,
  colframe=black,
  boxrule=0.6pt,
  arc=2pt,
  left=6pt,right=6pt,top=6pt,bottom=6pt,
  before skip=6pt, after skip=6pt
]

You are a medical expert answering domain-specific questions.\newline

    Your goal is to take small, reasoning that move toward one single correct answer.\newline

    OUTPUT CONTRACT\newline
    - You can have multiple arounds. But for each round, you must output EITHER:\newline
    (A) exactly ONE step line, OR\newline
    (B) exactly ONE answer line.\newline
    - A step line must start with <|step|> and contain ONE justification (20 - 30 words).\newline
    - An answer line must start with <|answer|> and contain ONLY the final answer (no explanation).\newline
    - For multiple choice questions, the final answer must be the option letter ONLY.\newline
    - Output <|answer|> if you are 100\% certain.\newline
    
    STRICT RULES\newline
    - Do NOT repeat earlier steps.\newline
    - Do NOT assume missing patient facts.\newline
    - Never use any tags other than <|step|> or <|answer|>.\newline

    BAD STEPS (do NOT do this)\newline
    - Multiple justifications in one step\newline
    - Repeat earlier steps\newline
    - Any tag like <|final\_answer|> (use <|answer|> only)\newline

\end{tcolorbox}
\caption{Inference Start Prompt}
\label{tab:inf_start_prompt}
\end{table*}

\begin{table*}[t]
\centering
\small
\begin{tcolorbox}[
  enhanced,
  breakable,
  title={Inference New Step Prompt},
  colback=white,
  colframe=black,
  boxrule=0.6pt,
  arc=2pt,
  left=6pt,right=6pt,top=6pt,bottom=6pt,
  before skip=6pt, after skip=6pt
]

If certain, output exactly ONE answer line\newline
    Otherwise, output exactly ONE step line\newline

    - A step line must start with <|step|> and contain ONE justification (20 - 30 words).\newline
    - An answer line must start with <|answer|> and contain ONLY the final answer (no explanation).\newline
    - For multiple choice questions, the final answer must be the option letter ONLY.\newline
    - Output <|answer|> if you are 100\% certain.\newline

    STRICT RULES\newline
    - Do NOT repeat earlier steps.\newline
    - Do NOT assume missing patient facts.\newline
    - Never use any tags other than <|step|> or <|answer|>.\newline

\end{tcolorbox}
\caption{Inference Step Prompt}
\label{tab:inf_step_prompt}
\end{table*}

\begin{table*}[t]
\centering
\small
\begin{tcolorbox}[
  enhanced,
  breakable,
  title={Inference Answer Prompt},
  colback=white,
  colframe=black,
  boxrule=0.6pt,
  arc=2pt,
  left=6pt,right=6pt,top=6pt,bottom=6pt,
  before skip=6pt, after skip=6pt
]

You MUST output exactly ONE line starting with <|answer|> and contain ONLY the final answer (no explanation).\newline

    - MUST 100\% START WITH <|answer|> \newline
    - Give the most likely answer if uncertain\newline
    - Avoid requesting more user inputs\newline
    - For multiple choice questions, the final answer must be the option letter ONLY.\newline

\end{tcolorbox}
\caption{Inference Answer Prompt}
\label{tab:inf_ans_prompt}
\end{table*}

\end{document}